\documentclass[journal]{IEEEtran}

\usepackage{graphics} 
\usepackage{epsfig} 
\usepackage{times} 
\usepackage{amsmath} 

\DeclareMathOperator*{\argmin}{arg\,min}
\usepackage[hidelinks]{hyperref}
\usepackage{amssymb}  
\usepackage{upgreek}
\usepackage{cite}
\usepackage{cleveref}
\usepackage{graphicx}
\usepackage{relsize}
\usepackage{longtable}
\usepackage{color,soul}
\usepackage[caption=false]{subfig}
\usepackage{multirow}
\usepackage{gensymb}
\usepackage{xcolor}

\usepackage{amsfonts}
\usepackage{algorithmic}
\usepackage{array}
\def\BibTeX{{\rm B\kern-.05em{\sc i\kern-.025em b}\kern-.08em
    T\kern-.1667em\lower.7ex\hbox{E}\kern-.125emX}}
\usepackage{balance}
\begin{document}

\title{Nonlinear Model Predictive Control with Obstacle Avoidance Constraints for Autonomous Navigation in a Canal Environment}
\author{Changyu Lee, Dongha Chung, Jonghwi Kim, and Jinwhan Kim 
\thanks{This research was supported by AVIKUS corp., and by a grant from National R\&D Project "Development of an electric-powered car ferry and a roll-on/roll-off power supply system" funded by Ministry of Oceans and Fisheries, Korea(PMS4420). \textit{(Corresponding author: Jinwhan Kim)}}
\thanks{The authors are with the Department of Mechanical Engineering, Korea Advanced Institute of Science and Technology, Daejeon 34141, South Korea (e-mail: leeck@kaist.ac.kr; chungdongha@kaist.ac.kr; stkimjh@kaist.ac.kr; jinwhan@kaist.ac.kr).}}

\maketitle

\begin{abstract}
In this paper, we describe the development process of autonomous navigation capabilities of a small cruise boat operating in a canal environment and present the results of a field experiment conducted in the Pohang Canal, South Korea. 
Nonlinear model predictive control (NMPC) was used for the online trajectory planning and tracking control of the cruise boat in a narrow passage in the canal. To consider the nonlinear characteristics of boat dynamics, system identification was performed using experimental data from various test maneuvers, such as acceleration-deceleration and zigzag trials. To efficiently represent the obstacle structures in the canal environment, we parameterized the canal walls as line segments with point cloud data, captured by an onboard LiDAR sensor, and considered them as constraints for obstacle avoidance. 
The proposed method was implemented in a single NMPC layer, and its real-world performance was verified through experimental runs in the Pohang Canal.
\end{abstract}

\begin{IEEEkeywords}
Marine robotics, integrated planning and control.
\end{IEEEkeywords}

\section{Introduction} \label{sec:Introduction}
\IEEEPARstart{I}{n} the maritime domain, autonomous surface vehicles (ASVs) are attracting considerable attention. Many studies have been conducted to increase the autonomy of ASVs \cite{review1}.
To achieve full autonomy, more complex and challenging marine environments, such as narrow channels or canals, must be considered.
In such environments, more sophisticated local trajectory planning and tracking algorithms that can reliably detect and efficiently react to hazardous structures and objects nearby are required. 
However, the under-actuated nature of marine vehicles and the limited space in canal areas pose challenges to achieving these developments.

Many studies have been conducted on local trajectory planning and tracking for ASVs. Because of their simplicity, graph search and vector field-based algorithms, such as A* and potential field algorithms, are frequently used to generate collision-free paths \cite{park2017development, singh2018constrained}. Line-of-sight (LOS) guidance and proportional–integral–derivative (PID) control algorithms are also widely used as tracking control methods \cite{lekkas2013line, yu2021finite}.
With the recent development of computational capabilities and resources, model predictive control (MPC), which requires a lot of computation, has been widely applied to trajectory planning and tracking.
MPC predicts a vehicle's motion for a finite time to generate a feasible trajectory and calculate the control inputs simultaneously through optimization.
While the control policy is being optimized, various constraints on the state and control input of ASVs can be explicitly considered, including collision avoidance conditions and a vehicle's nonlinear dynamics.

\begin{figure*}[t]
    \centerline{\includegraphics[width=\linewidth]{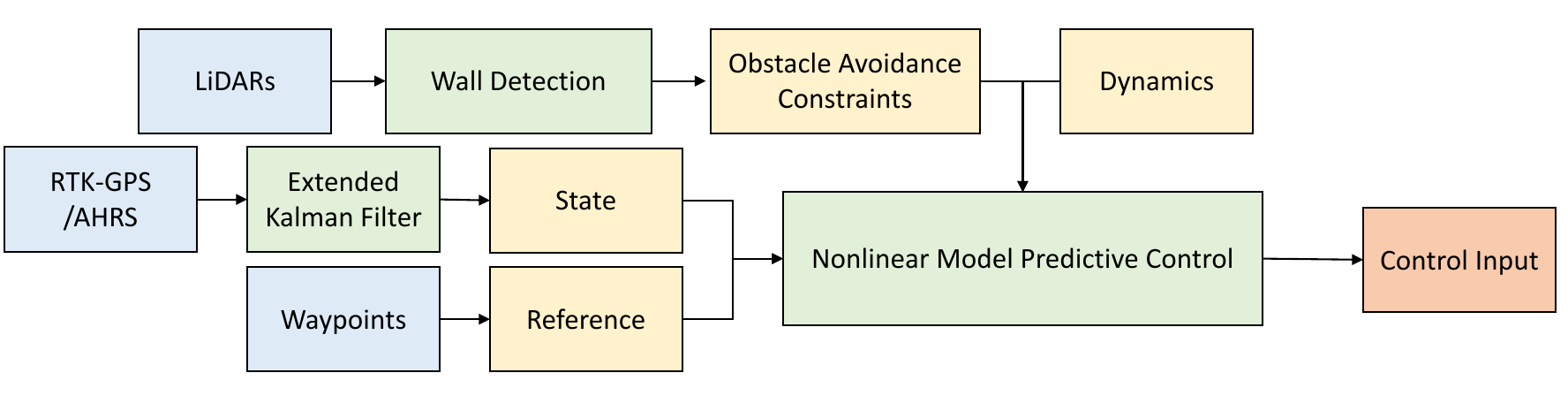}}
    \caption[Flowchart.]{Flowchart of the proposed algorithm.}
    \label{fig:flowchart}
\end{figure*}

However, there are many difficulties in applying the MPC-based algorithm to real-world applications because of the unknown vehicle dynamics and environment. For example, since MPC is a model-based control approach, a system identification process is desirable to estimate a reasonably accurate and reliable mathematical model of vehicle dynamics.
In addition, nearby obstacles must be detected in real-time with sufficiently high accuracy and reliability using onboard detection sensors.

For this reason, most studies have been conducted in simulation and laboratory environments where accurate obstacle's states and vehicle dynamics can be easily obtained.
In simulation studies, a cost function that minimizes path tracking error and energy consumption is usually used, and obstacle avoidance is considered through the circle, polygonal or elliptical constraints \cite{abdelaal2018nonlinear,bitar2019energy, abdelaal2021predictive,9483265,9482890}.
To verify the performance of the MPC in more realistic conditions, hardware-in-the-loop tests were performed in \cite{kosch2021hardware}, and in \cite{kinjo2021trajectory}, system identification was performed using experimental data, and the tracking performance was verified through simulation with a fully actuated real-scale ship.
In \cite{wang2018design}, optimization-based system identification was performed to identify mathematical models, and the tracking performance was verified through path tracking experiments in both indoor and outdoor environments with a quarter-scale robotic boat called ”Roboat”. 
In \cite{wang2019roboat}, an experiment using Roboat was performed using LiDAR measurements. A collision-free path was calculated considering obstacle avoidance constraints by the path planning algorithm, and the MPC algorithm was used for accurate path tracking.

\begin{figure}[t]
    \centerline{\includegraphics[width=\linewidth]{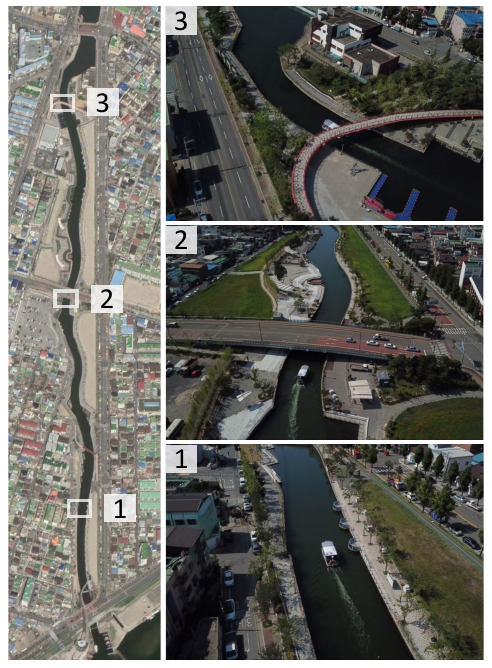}}
    \caption{Overview of the experimental site, Pohang Canal, which is located in Pohang, South Korea.}
    \label{fig:pohang_canal}
\end{figure}

In this paper, we developed an NMPC-based autonomous navigation algorithm in a canal environment.
To determine the nonlinear characteristics of vehicle dynamics, we performed optimization-based system identification using acceleration-deceleration and zigzag maneuvering data obtained in real operating conditions. 
We used three onboard LiDARs to detect and parameterize the obstacle structures in canal environment as line segments. This approach allows us to generalize the representation of any object shape as a combination of line segments.
The detected line segments were then used as obstacle avoidance constraints of the NMPC algorithm. 
This ensured that the identified nonlinear dynamics were considered in the implementation of obstacle avoidance, local trajectory planning, and tracking algorithms, which were integrated into a single NMPC optimization problem.
The overall framework of the proposed approach is illustrated in Fig.~\ref{fig:flowchart}. 
To validate the effectiveness of the proposed algorithm, we conducted simulations and field experiments using a 12-person cruise boat navigating through the 1 km-long Pohang Canal, which has an average width of 15 m (see Fig.~\ref{fig:pohang_canal}). 
To the best of our knowledge, our research represents the first attempt to autonomously navigate in a real canal environment with a full-size boat. 
The main contributions of this study can be summarized as follows:
\begin{itemize}
\item We propose a novel approach for autonomous navigation of an ASV in a canal environment using NMPC, which parameterizes nearby objects as a combination of line segments for obstacle avoidance constraints.
\item The proposed NMPC integrates trajectory planning, tracking control, and object detection algorithms to enhance the overall performance and safety of the ASV by ensuring the satisfaction of the obstacle avoidance constraints.
\item We validate the effectiveness of the proposed approach through simulations and real-world field experiments using a full-size cruise boat in the Pohang Canal.
\end{itemize}

\begin{figure}[t]
    \centering
    \includegraphics[width=0.9\linewidth]{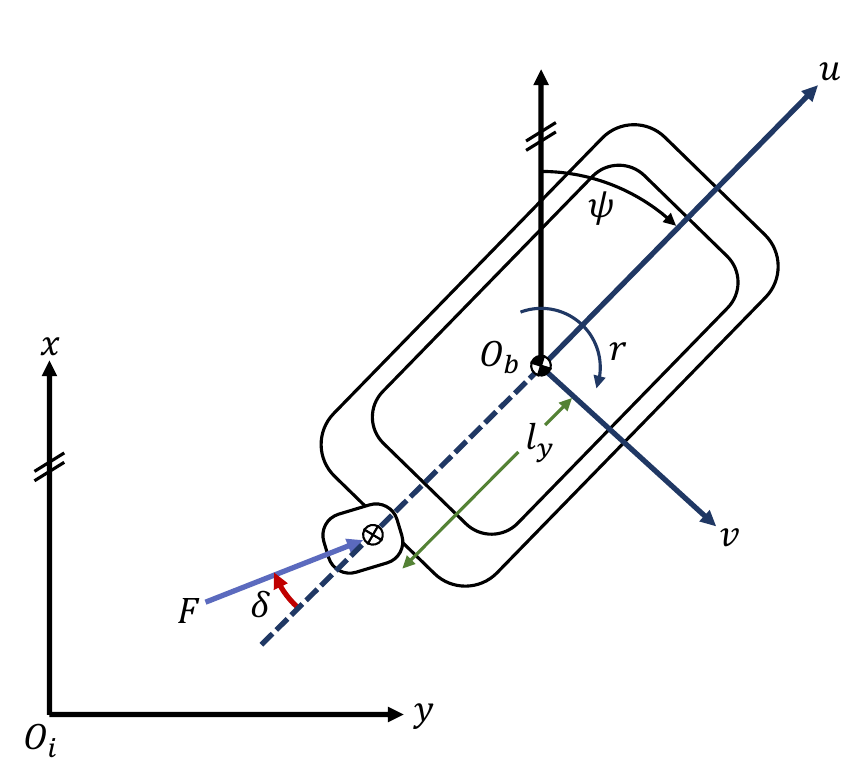}
    \caption[Coordinate system.]{Coordinate systems of the boat. The $O_b$ and $O_i$ represent the body-fixed and inertial coordinates, respectively, and $l_y$ is the distance to the motor.}
    \label{fig:coordinate}
\end{figure}

The following section presents a dynamic model of a boat and the procedure for system identification. Section~\ref{section3} presents the formulation of the proposed scheme, which includes LiDAR-based detection and NMPC algorithms. Section~\ref{section4} describes the results of system identification and the autonomous navigation experiments in the Pohang Canal. The conclusions of this study are presented in Section~\ref{section5}. 

\section{Vehicle Dynamics Modeling} \label{section2}
The dynamic model of a surface vehicle comprises kinematic and kinetic equations. These equations can be defined in the body-fixed and inertial coordinate systems, as shown in Fig.~\ref{fig:coordinate}. 
The following 3-DOF horizontal plane model was used:
\begin{subequations}   \label{eq:dynamics}%
   \begin{gather}
    M\dot{\nu}+C(\nu )\nu +D(\nu)\nu = \tau_{c} \label{eq:kinetics}\\ 
    \dot{\eta} = R(\psi)\nu \label{eq:kinematics} \\
    R(\psi) = \begin{bmatrix}
    \cos \psi  & -\sin\psi & 0 \\
    \sin\psi & \cos\psi & 0 \\
    0 & 0 &  1
    \end{bmatrix}
    \label{eq:rot}
   \end{gather}
\end{subequations}
where $ \nu = [u,\,v,\,r]^\top$ and $\eta = [x,\,y,\,\psi]^\top$ are the velocity and position vectors, respectively. $M$ is the inertia matrix, $C(\nu)$ is the Coriolis-centripetal matrix, $D(\nu)$ is the damping matrix, $\tau_c = [\tau_X,\,\tau_Y,\,\tau_N]^\top$ represents the control forces and moment in each direction, and $R(\psi)$ is the rotation matrix, which converts from the body-fixed coordinate to the inertial coordinate.

Assuming that the vehicle is symmetric in the $x$ and $y$ directions, $M$, $C(\nu)$, and 
$D(\nu)$ can be expressed as follows:
\begin{equation}
        M = \begin{bmatrix}  
    m_{11} & 0 & 0 \\
    0 & m_{22} & 0 \\
    0 & 0 & m_{33}
    \end{bmatrix}
    \label{eq:M_matrix} 
\end{equation}
\begin{equation}
C(\nu) = \begin{bmatrix} 
        0 & 0 & -m_{22}v \\
        0 & 0 &  m_{11}u \\
        m_{22}v & -m_{11}u & 0
\end{bmatrix}    \label{eq:C_matrix}
\end{equation}
\begin{equation}
D(\nu) = -\begin{bmatrix} 
X_u& 0 & 0 \\
0 & Y_v & Y_r \\
0 & N_v& N_r 
\end{bmatrix}
- \begin{bmatrix} 
X_{u|u|}|u| & 0 & 0 \\
0 & Y_{v|v|}|v| & Y_{r|r|}|r| \\
0 & N_{v|v|}|v| & N_{r|r|}|r|
\end{bmatrix}
\label{eq:D_matrix}
\end{equation}
where $m_{11}$, $m_{22}$, and $m_{33}$ are the mass and moments of inertia, including the added mass and added moment of inertia. $X_u$, $Y_v$, $Y_r$, $N_v$, and $N_r$ are the linear drag coefficients and $X_{u|u|}$, $Y_{v|v|}$, $Y_{r|r|}$, $N_{v|v|}$, and $N_{r|r|}$ are the nonlinear drag coefficients.

\begin{figure}[t]
    \centerline{\includegraphics[width=\linewidth, trim={0 0 0 -0.5cm},clip]{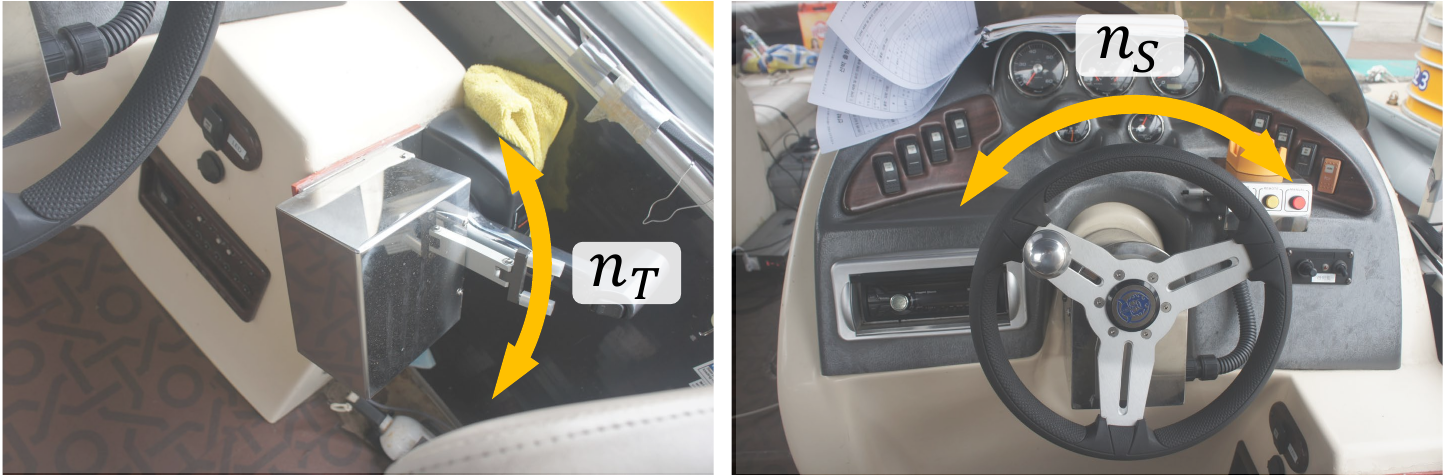}}
    \caption[Control devices.]{Control devices used for the vehicle.}
    \label{fig:devices}
\end{figure}

\begin{figure}[t]
    \centering
    \includegraphics[width=\linewidth]{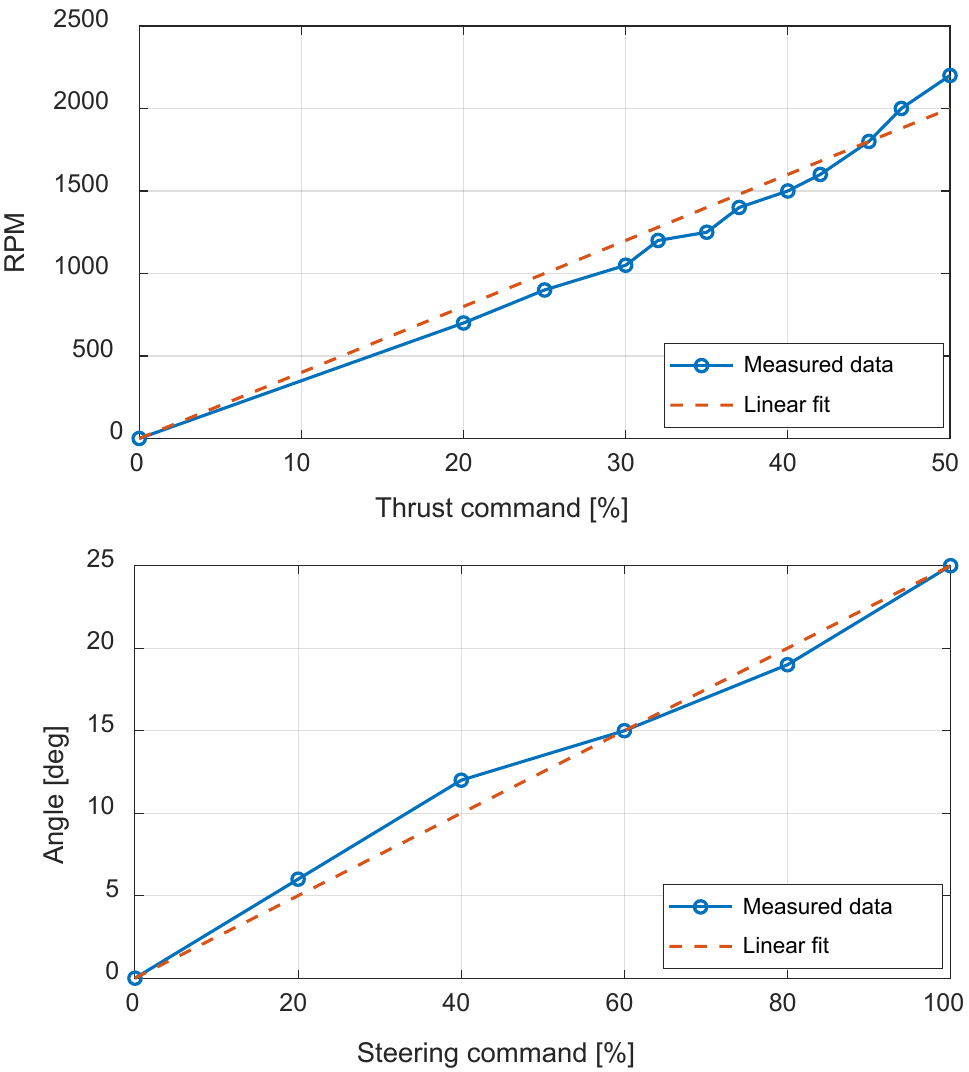}
    \caption[Linear fit of control command data.]{Linear fit of the control command data.}
    \label{fig:mapping}
\end{figure}

The control force and moment $\tau_c$ are a function of the propeller rotational speed $n$ and the angle of the outboard motor $\delta$. As shown in Fig.~\ref{fig:devices}, we controlled the boat with the throttle $n_T$ and the steering wheel $n_S$. These values were measured from -100 to 100\%. To ascertain the relationship between $[n_T,\, n_S]^\top$ and $[n,\, \delta]^\top$, and the data are shown in Fig.~\ref{fig:mapping}. Through these values, it can be assumed that these variables have a linear relationship with one another. Finally, based on the fact that thrust is proportional to the square of the propeller rotation speed, $[\tau_X, \,\tau_Y, \,\tau_N]^\top$ can be expressed as follows:
\begin{subequations}\label{eq:force}
   \begin{gather}
    \begin{bmatrix}
    \tau_X \\
    \tau_Y \\
    \tau_N
    \end{bmatrix} =
    \begin{bmatrix}
    F \cos \delta \\
    F \sin \delta \\
    - l_y F \sin \delta
    \end{bmatrix} =
    \begin{bmatrix}
    c n_T^2 \cos (\alpha n_{S}) \\
    c n_T^2 \sin (\alpha n_{S}) \\
    - l_y c n_T^2 \sin (\alpha n_{S}) 
    \end{bmatrix} \\
    \alpha = \frac{\delta_{\text{max}}}{100}
   \end{gather}
\end{subequations}
where $c$ is the unknown control coefficient, $l_y$ is the distance from the center of the body-fixed coordinate to the outboard motor, and $\delta_{\text{max}}$ is the maximum angle of the outboard motor. The complete dynamic equation can be reformulated by combining \eqref{eq:dynamics}-\eqref{eq:force} as follows:
\begin{equation}
\dot{\mathbf{x}} = f(\mathbf{x},\mathbf{u}, P)
\label{eq:final}
\end{equation}
where $\mathbf{x} = [x,\,y,\,\psi,\,u,\,v,\,r]^\top$ and $\mathbf{u} = [n_T,\,n_S]^\top$ are the state and control input vectors, respectively, and $ P = [c,\, m_{11},\, m_{22},\, m_{33},\, X_u,\, Y_v,\, Y_r,\,N_v,\, N_r,\, X_{u|u|},\, Y_{v|v|},\, Y_{r|r|}, \\ \, N_{v|v|},\, N_{r|r|}]$ is the set of unknown parameters.

\begin{figure*}[t]
    \centering
    \begin{minipage}{0.31\linewidth}
            \subfloat[][]
            {\includegraphics[width=\textwidth]{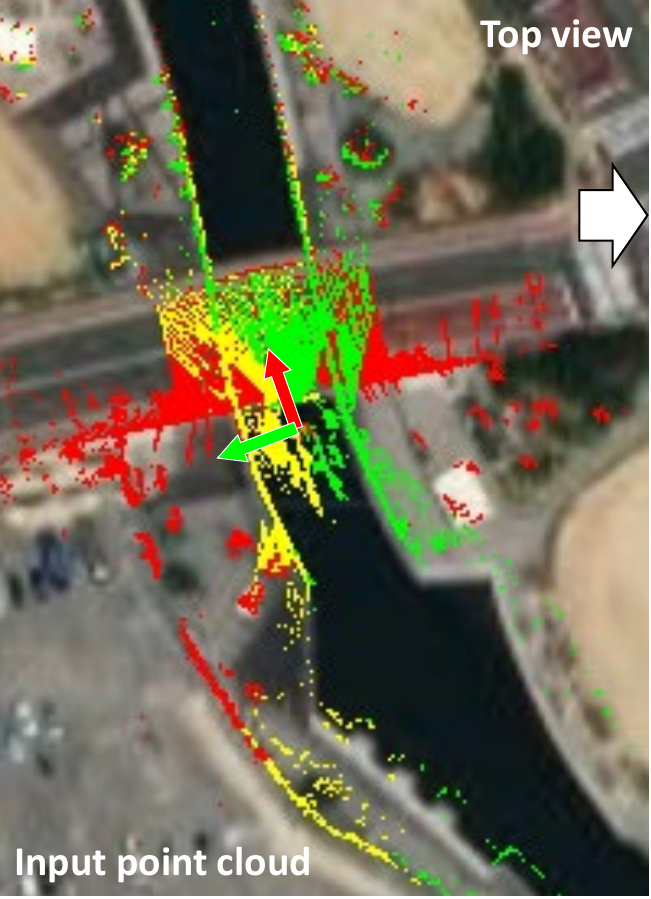}}
    \end{minipage}
    \begin{minipage}{0.29\linewidth}
            \subfloat[][]
            {\includegraphics[width=\textwidth]{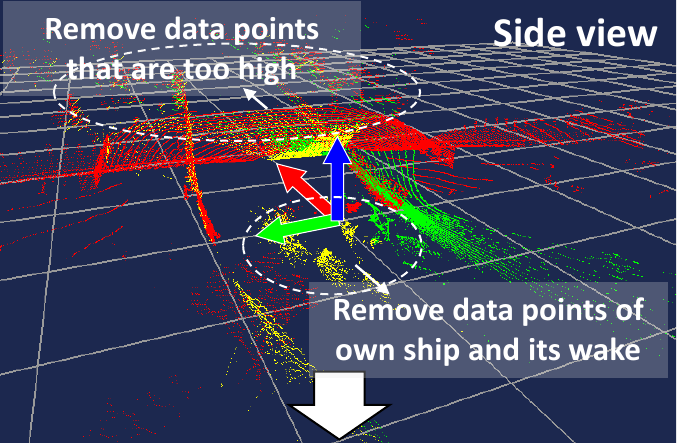}}\\
            \subfloat[][]
            {\includegraphics[width=\textwidth]{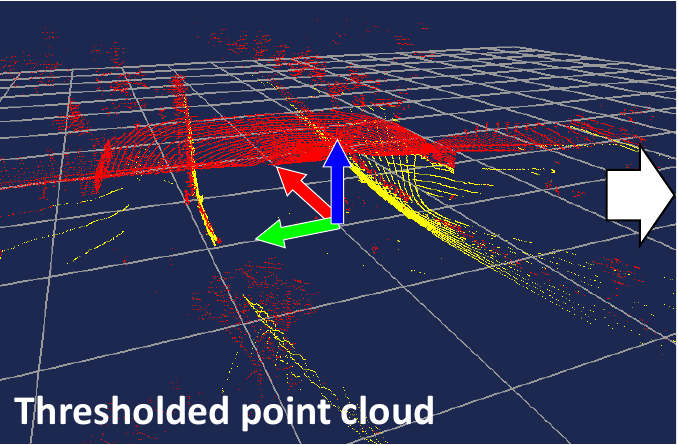}}
    \end{minipage}
    \begin{minipage}{0.31\linewidth}
            \subfloat[][]
            {\includegraphics[width=\textwidth]{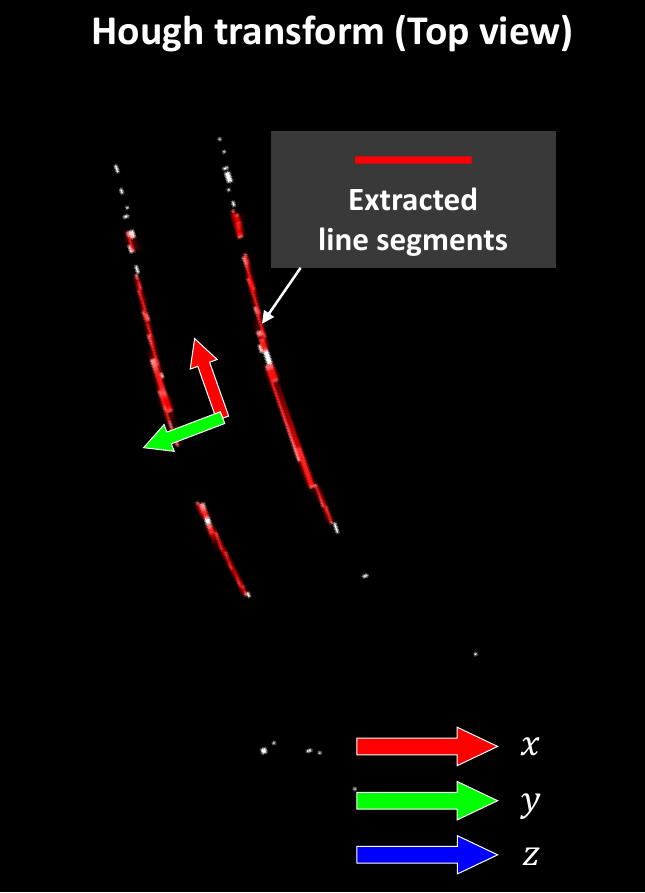}}
    \end{minipage}
     \caption{Visualization of the LiDAR-based obstacle detection algorithm. The input point cloud, obtained from three LiDARs, is depicted in (a) as a top-view and in (b) as a side view. To remove unnecessary point cloud data, we defined threshold limit in the $x$, $y$, and $z$ directions. The resulting data is shown in (c). Lastly, we utilized the Hough transform to extract line segments, which are depicted in a top-view in (d).} 
    \label{fig:lidar}
\end{figure*}

For the controller design, the unknown parameters of the dynamic equation in \eqref{eq:final} must be identified. To determine these parameters, the nonlinear programming (NLP) problem can be posed as follows:
\begin{subequations}
\label{eq:SI} %
\begin{gather}
{P^*} = \argmin_{P} \sum_{i=0}^{N}
    (\mathbf{x}_i -\bar{\mathbf{x}}_i)^\top W (\mathbf{x}_i-\bar{\mathbf{x}}_i) 
    \label{eq:SI_optim}
\end{gather}
\begin{align}
    \text{s.t. } \mathbf{x}_{i+1}  &= f_d(\mathbf{x}_{i},\bar{\mathbf{u}}_{i}, P) \\
    \mathbf{x}_0  &= \mathbf{x}_{init} 
    \label{eq:SIconstraints}
\end{align}
\end{subequations}
where $N$ is the number of data samples, $W$ is the state weight matrix, $\bar{\mathbf{x}}$ and $\bar{\mathbf{u}}$ are the state and control input of the experimental data, respectively, $f_d$ is the discretized system of $f$ in \eqref{eq:final} with a sampling time of the dataset, and $\mathbf{x}_{init}$ is the initial state. 

The formulated NLP in \eqref{eq:SI} contains a large number of variables and constraints. Therefore, we adapted the interior point algorithm for this NLP, which is known to be effective for large-scale problems. We used the IPOPT \cite{IPOPT} in CasADI \cite{casadi} in the MATLAB environment. 
Explanations of the dataset, initial guess, and constraints are provided in Section \ref{section4}. 

\section{Trajectory Planning and Control} \label{section3}
\subsection{LiDAR-based obstacle detection algorithm}
Three sets of point clouds from three LiDARs were merged to detect any objects around the vehicle. To detect only the obstacles in a waterway, the onshore structures, such as buildings and paved roads, were excluded by thresholding the point cloud in the $z$-direction, and wakes were excluded by setting the thresholds in the $x$ and $y$ directions. A 2D occupancy grid map was generated by projecting the remaining points onto the horizontal plane. The Hough transform \cite{hough1962method} was used to detect the line segments of the objects. The detected $i$-th line segment is expressed by the center position $p_i = (x_{c,i},\,y_{c,i})$, angle $\theta_i$, and length $l_i$ as follows:
\begin{equation}
    L_i(p_i,\theta_i,l_i).
    \label{eq:line}
\end{equation}
The flow of the detection algorithm is shown in Fig.~\ref{fig:lidar}.

\subsection{Nonlinear model predictive control}

To avoid the detected obstacles and follow the predefined path with minimum input effort while considering the dynamic characteristics of the boat, we formulated the optimal control problem. Redefine the state and input vector as $\mathbf{x} = [x,\,y,\,\psi,\,u,\,v,\,r,\,n_T,\,n_s]^\top$ and $\mathbf{u} = [\Delta n_T,\,\Delta n_S]^\top$, respectively.
Using a multiple shooting scheme, we designed a discrete cost function as follows:
\begin{subequations}
\label{NMPC} %
\begin{equation}
\min_{\mathbf{x}(\cdot),{\mathbf{u}}(\cdot)}\sum_{i=0}^{N_p-1}
\ell (\mathbf{x}_i, \mathbf{r}_i, \mathbf{u}_i, s_i) + \ell_T(\mathbf{x}_{N_p},\mathbf{r}_{N_p}, s_{N_p})
\end{equation}
\begin{align}
    \text{s.t. } \mathbf{x}_0 - \mathbf{x}_{init} &= 0,  \label{mpc:init} \\[5pt]
    \mathbf{x}_{i+1} - f_d(\mathbf{x}_i,\mathbf{u}_i,P^*) &= 0 , \ i = 0,\ldots,N_p-1, \label{mpc:dynamics} \\[5pt]
    g(\mathbf{u}_i) & \leq 0, \ i=0,\ldots,N_p, \label{mpc:constraints1} \\[5pt]
    h(\mathbf{x}_i, L_j) & \leq 0, \ i=0,\ldots,N_p, \ j=0,\ldots,N_l, \label{mpc:constraints2} 
\end{align}
\end{subequations}
where $ \mathbf{r}_i = [x_{r,i},\,y_{r,i},\,\psi_{r,i},\,u_{r,i},\,0,\,0,\,0,\,0]^\top $ is the reference state, $\mathbf{x}_{init}$ is the initial state, $s_i$ is the slack variable, $\ell$ is the stage cost function, $\ell_T$ is the terminal cost function, $N_p$ is the prediction horizon, and $P^*$ is the estimated set of parameters obtained by solving \eqref{eq:SI}. \eqref{mpc:constraints1} and \eqref{mpc:constraints2} are the inequality constraints for the control input and obstacle avoidance, respectively, and $N_l$ is the number of line segments.

The stage and terminal cost functions penalize the error between the predicted states and reference states as follows:
\begin{subequations}
\begin{align}
\ell(\mathbf{x}_i, \mathbf{r}_i, \mathbf{u}_i, s_{i}) &= (\mathbf{x}_i - \mathbf{r}_i)^\top Q (\mathbf{x}_i - \mathbf{r}_i) + \mathbf{u}_i^\top R \mathbf{u}_i + \rho s_i^2  \\
\ell_T(\mathbf{x}_{N_p},\mathbf{r}_{N_p}, s_{N_p}) &= (\mathbf{x}_{N_p} - \mathbf{r}_{N_p})^\top Q_T (\mathbf{x}_{N_p} - \mathbf{r}_{N_p})+ \rho s_{N_p}^2
\end{align}
\end{subequations}
where the matrices $Q$, $R$, and $Q_T$ represent the weight matrices of the cost function that penalizes the state error, rate of change of control input, and terminal state error, respectively. $\rho$ is the weighting factor for penalizing the slack variables. 

The reference states were defined with predefined path based on the current waypoint $(a_1,a_2)$ and the next waypoint $(b_1,b_2)$, with the following equations:
\begin{equation}
\begin{bmatrix}
x_{r,i+1} \\
y_{r,i+1} \\
\end{bmatrix}
= \begin{bmatrix}
x_{r,i} \\
y_{r,i} \\
\end{bmatrix} + 
\begin{bmatrix}
u_{r,i} \cos \psi_{r,i} \\
u_{r,i} \sin \psi_{r,i} \\ 
\end{bmatrix}T_s
\end{equation}
where $u_{r,i}$ is the target speed and was determined based on the normal operating speed of the cruise boat in the Pohang Canal.  
The reference position $(x_{r,i}, y_{r,i})$ is determined by the closest point on the waypoint path from the current state.
$T_s$ represents the prediction sampling time and the reference heading angle $\psi_{r,i}$ for the current path was determined using the following equation:
\begin{equation}
    \psi_{r,i} = \arctan \left(\frac{b_2 - a_2}{b_1 - a_1}\right).
\end{equation}

\begin{figure}[t]
    \centerline{\includegraphics[width=\linewidth, trim={0 0 0 -0.5cm},clip]{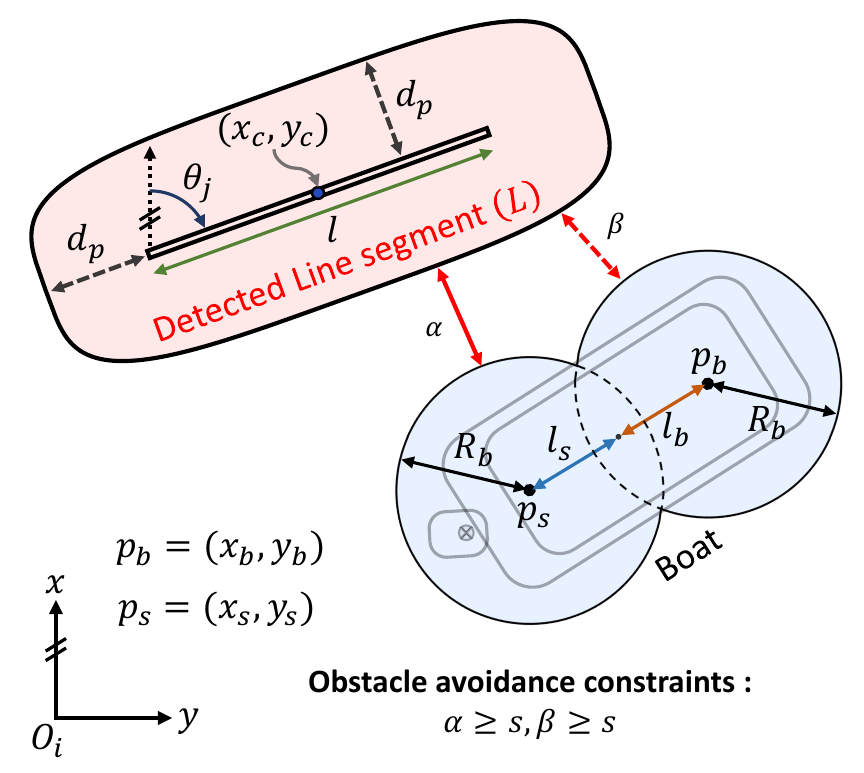}}
    \caption[Inequality constraints for obstacle avoidance.]{Illustration of the geometric constraints for safe separation. Here, $\alpha = d(x_b,y_b,L)-R_b-d_p$ and $\beta = d(x_s,y_s,L)-R_b-d_p$, respectively.}
    \label{collision}
\end{figure}

\begin{figure}[t]
    \centerline{\includegraphics[width=\linewidth]{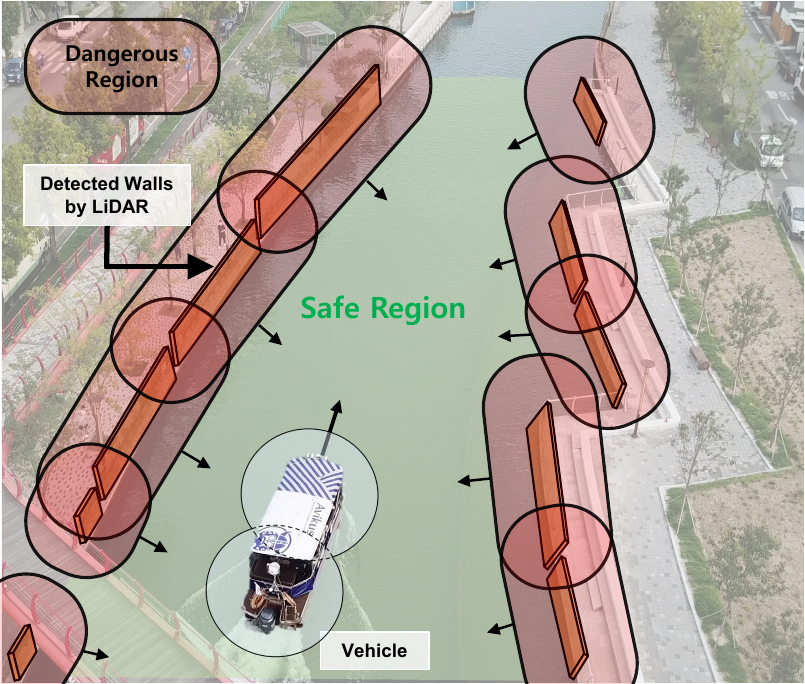}}
    \caption[Inequality constraints for obstacle avoidance.]{Illustration of the obstacle avoidance constraints.}
    \label{collision_detail}
\end{figure}

The inequality constraints for the control inputs and their rate of change in \eqref{mpc:constraints1} were defined as follows:
\begin{equation}
\begin{aligned}
     | n_T | \leq n_{T,\text{max}}&, \ | n_S | \leq n_{S,\text{max}}\\ 
    | \Delta{n_T} | \leq \Delta{n_{T,\text{max}}}&, \ 
    | \Delta{n_S} | \leq \Delta{n_{S,\text{max}}}  
    \label{mpc:constraints}
\end{aligned}
\end{equation}
where subscript $(\cdot)_{\text{max}}$ indicates the maximum of the corresponding variables. 
The inequality constraints for obstacle avoidance \eqref{mpc:constraints2} were defined as follows:
\begin{equation}
\begin{aligned}
     d(x_{b,i}, y_{b,i},L_j) &\geq R_{b} + d_p + s_i\\
     d(x_{s,i}, y_{s,i},L_j) &\geq R_{b} + d_p + s_i 
\label{eq:soft_collision}
\end{aligned}
\end{equation}
where $p_{b,i} = (x_{b,i}, y_{b,i})$ and $p_{s,i}= (x_{s,i}, y_{s,i})$ indicate the center positions of two circles of radius $R_{b}$ representing the safety boundary of the boat as shown in Figs.~\ref{collision} and \ref{collision_detail}, where $l_s, l_b = 2$ m. The function $d(x, y, L)$ indicates the distance between the position $(x, y)$ and the detected line segments $L$. $d_p$ is the desired separation, and the slack variable $s_i$ was introduced to make it a soft constraint to allow a slight violation of safe separation. The constraints can be approximated by the following differentiable function:

\begin{multline}
    \bigg(\frac{(x-x_c)\cos\theta +(y-y_c)\sin\theta }{l/2 + R_b + d_p }\bigg)^4 +  \\ \bigg(\frac{-(x-x_c)\sin\theta+(y-y_c)\cos\theta}{R_b + d_p + s_i}\bigg)^4 - 1 \geq 0.
    \label{mpc:pnorm}
\end{multline} 
Figure~\ref{collision_detail} illustrates the collision avoidance constraint. It defines a dangerous region where the vehicle's distance from the line segment is less than the prescribed safety distance. Any position inside this region violates the obstacle avoidance constraint, whereas remaining within the safe area guarantees continuous compliance with the constraint, thereby enhancing safe navigation.


The real-time iteration algorithm \cite{houska2011acado}, generated by the ACADO Code Generation Toolkit, was used to solve the real-time NMPC problem formulated in \eqref{NMPC}.
The NLP was calculated by using an SQP algorithm, and the quadratic program was solved using a parametric active-set algorithm \cite{ferreau2014qpoases}.

\section{Simulation and Experimental Results} \label{section4}
The proposed approach's effectiveness was validated through numerical simulation and real-world experiments in this study. Initially, the system model was identified using experimental data, and simulations were conducted utilizing the identified ship dynamic model. Subsequently, real-world experiments were carried out in the Pohang Canal, and the results were analyzed and discussed.

\subsection{System identification results}
\begin{figure}[t]
    \centerline{\includegraphics[width=\linewidth, trim={0 0 0 -0.5cm},clip]{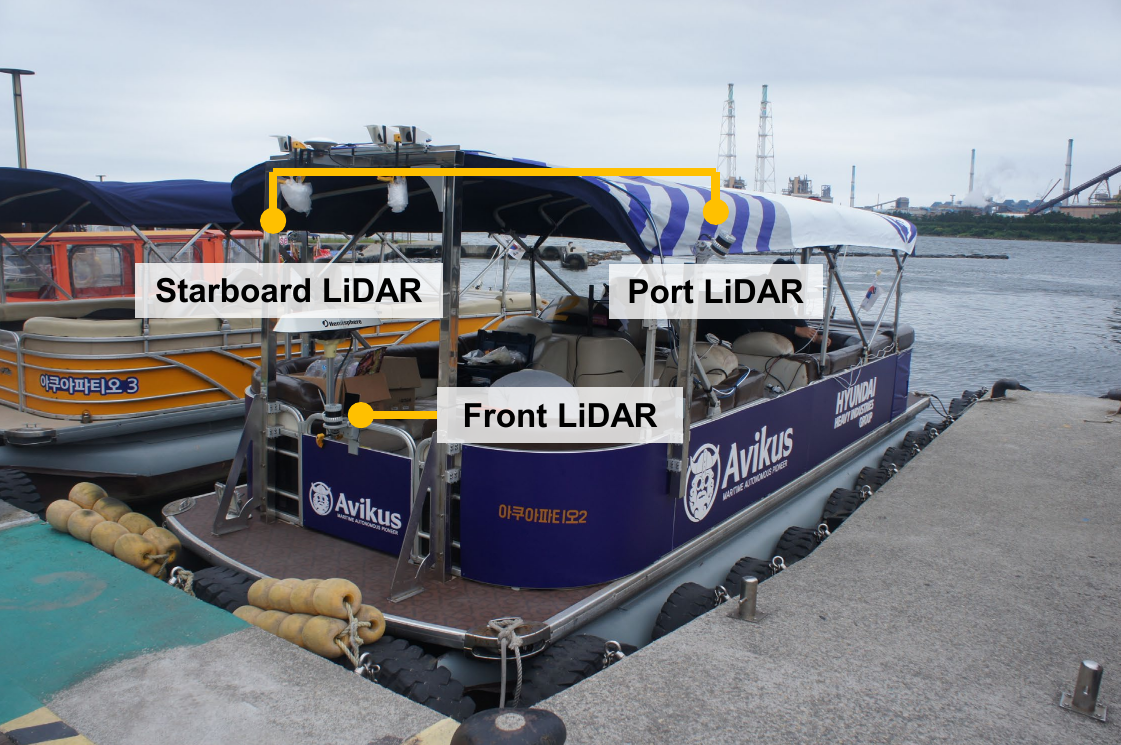}}
    \caption[AquaPatio.]{The cruise boat and 3D LiDARs used for the experiment.}
    \label{fig:aquapatio}
\end{figure}

\begin{figure}[t]
    \centering
    \begin{minipage}{\linewidth}
        \centering
        \subfloat[][]
        {\includegraphics[width=\linewidth]{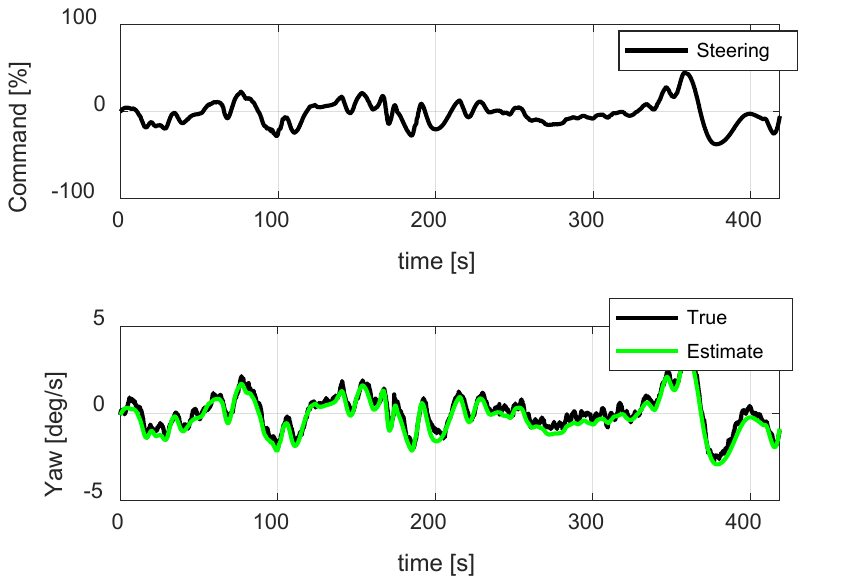}}
    \end{minipage}
    \begin{minipage}{\linewidth}
        \centering
        \subfloat[][]
        {\includegraphics[width=\linewidth]{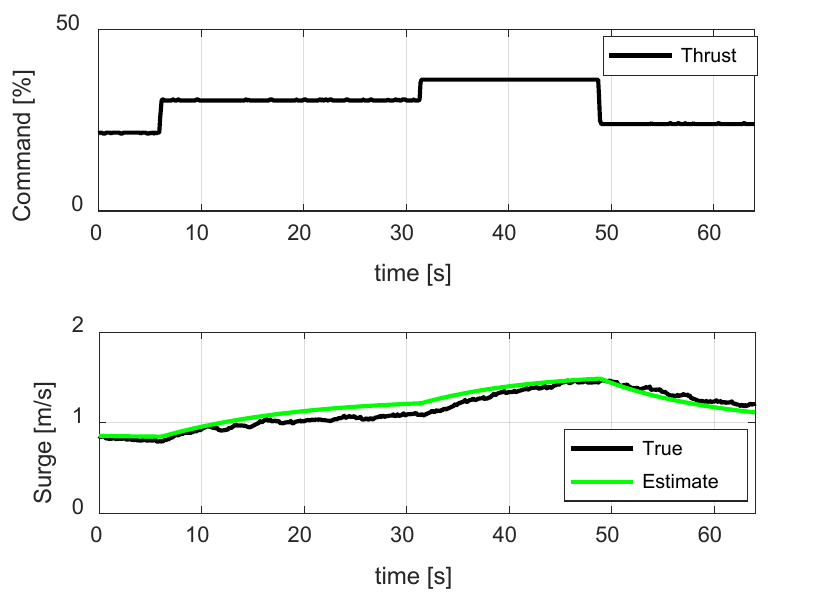}}
    \end{minipage}
    \caption{System identification results. The black and green lines indicate the measured data and the estimated data, respectively. In (a), $n_S$ was maintained at 0\%, and in (b), $n_T$ was maintained at 42.0\%. }
    \label{fig:zigzag}
\end{figure}

To identify the system, data were gathered in an open water area next to the Pohang Canal.
Acceleration-deceleration and zigzag maneuvering experiments were performed. In the acceleration and deceleration tests, an arbitrary thrust command set was used.
This thrust command set is described in Table~\ref{tab:SI_test}. Each command was held until the boat reached the steady-state.
The zigzag test was performed for 20$\degree$/50\% conditions. 
A $\pm 50\%$ steering command was given under the constant speed condition when the heading was equal to $\mp 20\degree$.
The zigzag test was ceased after three overshoots.
To simplify the problem with many unknown parameters, we decoupled the surge and sway-yaw models, and the optimization was performed sequentially using two pieces of data by guessing the initial parameter values based on empirical methods.

\begin{table}[h]
\centering
\renewcommand{\arraystretch}{1.3}
\caption{System identification data} 
\label{tab:SI_test}
\begin{tabular}{cc}
\hline
\textbf{Test item} & \textbf{Description} \\
\hline
\hline
Acceleration & $n_T$ = 31.0, 34.9, 38.6, 41.0, 50.6 $\%$ \\
Deceleration & $n_T$ = 50.6, 39.4, 20.0 $\%$ \\
\hline
Zigzag & 20$\degree$/$50 \%$ test, $n_T$ = $42.0 \%$ \\
\hline
\end{tabular}
\end{table}

\begin{table}[t]
\caption {System identification results} 
\centering
\renewcommand{\arraystretch}{1.3}
\begin{tabular}{c c | c c}
\hline \textbf{Item} & \textbf{estimated value} & \textbf{Item} & \textbf{estimated value} \\
\hline 
\hline 
$m_{11}$ & 1.9149e+03   & $N_r$    & -2.1940e+03 \\
$m_{22}$ & 1.8238e+03   & $X_{u|u|}$ & -54.344     \\
$m_{33}$ & 1.9351e+03   & $Y_{v|v|}$ & -282.62    \\
$X_u$    & -29.220     & $Y_{r|r|}$ & -0.0025      \\
$Y_v$    & -3.6284e+03  & $N_{v|v|}$ & -0.0010      \\
$Y_r$    & -1.6080e-04  & $N_{r|r|}$ & -206.44    \\
$N_v$    & -1.3102e-04  & $c$        & 1.3331e-05   \\
\hline
\label{tab:SIresult}
\end{tabular}
\end{table}

\begin{figure*}[t]
    \centering
    \includegraphics[width=\linewidth]{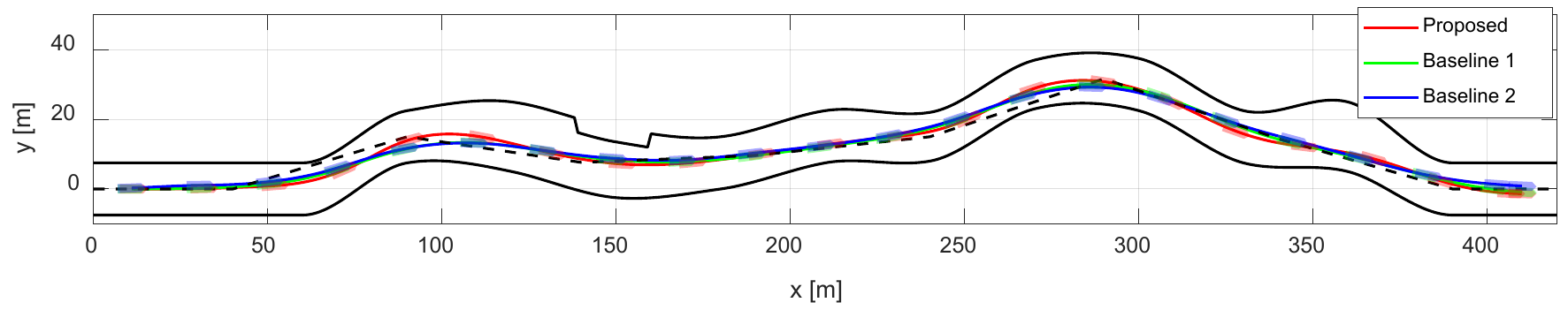}
    \caption{Simulation results: Time trajectories of the system states and inputs from three different algorithms are represented.}
    \label{fig:sota_sim}
\end{figure*}

Surge model identification was performed using the acceleration-deceleration data. The weight in air and the moment of inertia calculated by the main dimensions of the boat were used as the initial guess. The following weight matrix was used:
\begin{equation}
    W = \text{diag}([0,\,0,\,0,\,1,\,0,\,0]).
\end{equation}
To identify the sway-yaw model, the value estimated through surge model identification and the following weight matrix was used:
\begin{equation}
W = \text{diag}([1,\,1,\,100,\,0,\,100,\,100]).
\end{equation}

Verification was performed through the operational data of the boat in the canal. The results are shown in Fig.~\ref{fig:zigzag}. The black line represents the actual data, and the green line represents the estimated value based on the identified model. The estimated values of each unknown parameter are provided in Table~\ref{tab:SIresult}.

\begin{table}[h]
\centering
\renewcommand{\arraystretch}{1.3}
\caption {NMPC parameters} 
\label{tab:MPC_Setting}
\begin{tabular}{c | c | c | c}
\hline
\textbf{Symbol}&  \textbf{Value} & \textbf{Symbol}&  \textbf{Value} \\
\hline
\hline
$N_p$ & 25 & $T_s$ & 1.0 sec \\
${n_{T,\text{max}}}$ & 100\% & ${n_{S,\text{max}}}$ & 100\%\\
${\Delta n_{T,\text{max}}}$ & 10\%/s &  ${\Delta n_{S,\text{max}}}$ & 40\%/s\\
$R_{b}$ & 3.0 m & $\delta_{\text{max}}$ & 25\degree \\
$d_p$ (Simulation) & 2.0 m & $d_p$ (Experiment)& 4.0 m \\
\hline
\hline
\textbf{Symbol}&  \multicolumn{3}{c}{\textbf{Value}} \\
\hline
\hline
$Q$  & \multicolumn{3}{c}{$\text{diag}([1,\, 1,\, 500, \,10,\, 0,\, 1000,\, 0,\, 0])$} \\
$R$  & \multicolumn{3}{c}{$\text{diag}([0.0001,\, 0.0001])$} \\
$Q_T$  & \multicolumn{3}{c}{$Q  N_p$} \\
$\rho$ & \multicolumn{3}{c}{10000} \\
\hline
\end{tabular}
\end{table}

\subsection{Simulation results of trajectory planning and control}\label{sec:sim}

To evaluate the performance of our proposed NMPC-based approach for integrated path planning and control, we compared it with two state-of-the-art approaches \cite{villa2020path, shan2020receding} in a simulation environment using the identified model.
The control frequency was set to 10 Hz, and we assumed that the canal boundaries within 50 m could be precisely detected in the simulation environment. For the NMPC algorithm, we used a prediction time of 25 seconds and a sampling time of 1.0 seconds. The thrust and steering commands had values ranging from -100\% to 100\%, and the maximum change rates were 10\%/sec and 40\%/sec, respectively, determined by the actual speed of the control device. The boat radius, $R_{b}$, was set to 3.0 m, and the desired separation, $d_p$, was 2.0 m. Refer to Table~\ref{tab:MPC_Setting} for detailed parameter settings.

\begin{figure}[t]
    \centering
    \includegraphics[width=\linewidth]{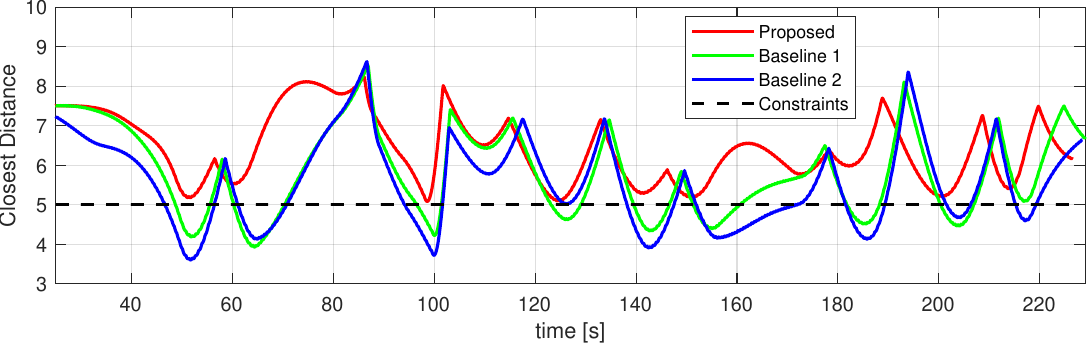}
    \caption{The separation distances by proposed and baseline algorithms.}
    \label{fig:sota_sim2}
\end{figure}

\begin{figure*}[t]
    \begin{minipage}{0.33\linewidth}
            \includegraphics[width=\linewidth]{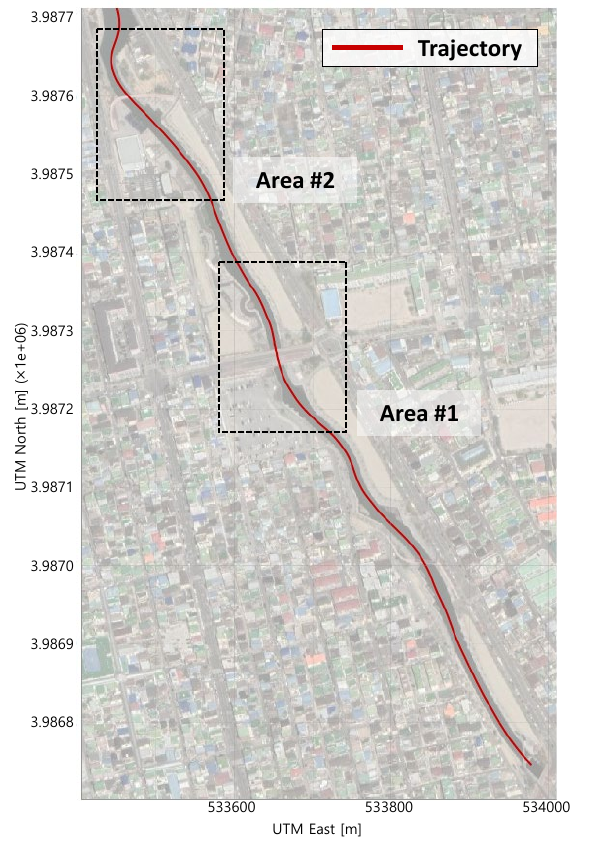}
    \end{minipage}
    \begin{minipage}{0.33\linewidth}
            \subfloat[][Close-up shot of area \#1 (Top view).]
            {\includegraphics[width=\linewidth]{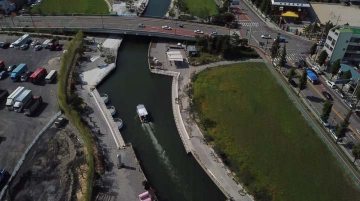}\label{fig:ca1}}\\
            \subfloat[][Close-up shot of area \#1 (Front camera view).]
            {\includegraphics[width=\linewidth]{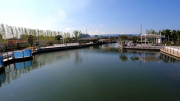}\label{fig:ca2}}        
    \end{minipage}
    \begin{minipage}{0.33\linewidth}
            \subfloat[][Close-up shot of area \#2 (Top view).]
            {\includegraphics[width=\linewidth]{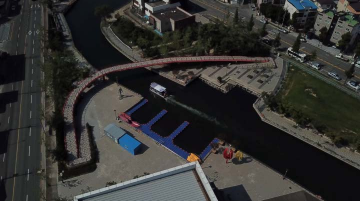}\label{fig:ca3}}\\
            \subfloat[][Close-up shot of area \#2 (Front camera view).]
            {\includegraphics[width=\linewidth]{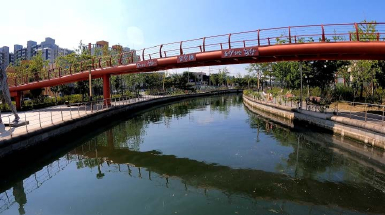}\label{fig:ca4}}
    \end{minipage}    
    \caption{The experimental trajectory plot on Google Maps. Close-up shot of each area in top and front camera views in (a)-(d).}
    \label{fig:trajectory}
\end{figure*}

The first baseline algorithm proposed in \cite{villa2020path} comprised two separate modules: a collision-free path planner and a tracking controller. In the first baseline algorithm, the path planning algorithm was designed to adaptively change the reference path by relocating waypoints appropriately, and the control algorithm was implemented by combining LOS guidance and PID control laws. For comparison simulation, the proposed NMPC algorithm was used for path planner.

The second baseline algorithm proposed in \cite{shan2020receding} used a receding horizon lexicographic path planning algorithm for surface vehicles in an urban waterway. Three costs (collision risk, heading variation, and distance) were sequentially optimized based on their priority.
The algorithm had a fixed endpoint condition and sampled candidate waypoints around a reference path. To apply the algorithm in a canal environment, we made modifications, which are detailed in the Appendix. 

In the simulation, we generated a canal environment and defined a reference path using a black dashed line, as shown in Fig.~\ref{fig:sota_sim}. The results of the state-of-the-art methods, represented by the green line from \cite{villa2020path} and the blue line from \cite{shan2020receding}, were compared against our proposed approach.
To evaluate each approach's obstacle avoidance capability, we used the closest distance metric, as depicted in Fig.~\ref{fig:sota_sim2}. At each state, we calculated the minimum distance to the canal boundary.
Our results show that our proposed approach outperforms the state-of-the-art methods in terms of meeting obstacle avoidance constraints, which is crucial for safe navigation in narrow waterways. Furthermore, the proposed algorithm has an average computation time of 0.0188 s and a maximum time of 0.0491 s, demonstrating that it can operate at a frequency of 10 Hz.

\subsection{Experimental Setup} 
In the experiment, we used the Robot Operating System for the communication between nodes.
To continuously track the pose of the boat, we designed a navigation filter by applying the extended Kalman filter framework using the sensor measurements from the attitude heading reference system (AHRS) and global positioning system (GPS). Each sensor delivered updated measurements at 100 Hz and 5 Hz. The state of the boat contained pose and linear velocity. 
In addition, three 3D LiDARs were used for detection as shown in Fig.~\ref{fig:aquapatio}. 
The front-facing LiDAR was located at the fore part of the boat, and its field of view was blocked by the boat’s own structure and limited to the front area. To cover the blind zone, the port and starboard LiDARs were additionally installed, slightly tilted downwards, to detect the sidewalls of the canal and nearby objects on both sides of the boat.
As a platform, we used a 12-person cruise boat operating in the Pohang Canal.
Detailed specifications of the boat are given in Table~\ref{tab:Model_Setting}.

\begin{table}[t]
\centering
\renewcommand{\arraystretch}{1.3}
\caption{Boat specifications} 
\label{tab:Model_Setting}
\begin{tabular}{cc}
\hline
\textbf{Item} & \textbf{Description} \\
\hline
\hline
\multicolumn{2}{l}
{\text{(a) Platform specification} }\\
\hline
Weight &  1,700 kg (in air)  \\
Length overall & 7.9 m  \\
Breadth &  2.6 m  \\
Draft  & 0.3 m \\
$l_y$  & 3 m \\
Propulsion  & Outboard gasoline engine\\
Power & 200 hp \\
Maximum speed  & $> 15$ knots \\
\hline
\hline 
\multicolumn{2}{l}
{\text{(b) Sensor specification}} \\
\hline
\multirow{2}{*}
{LiDAR (front)} & 3D, 0.5 m - 120 m range,\\[-3pt] & 64 vertical resolution \\
\multirow{2}{*}
{LiDAR (side)} & 3D, 0.5 m - 120 m range, \\[-3pt] & 32 vertical resolution\\
\multirow{2}{*}
{RTK GPS} & $\pm$0.01 m + 1 ppm CEP \\[-3pt] & RTK accuracy\\
\multirow{2}{*}
{AHRS} & 0.02 mg accelerometer resolution, \\[-3pt] & 0.003$\degree$/sec gyroscope resolution\\
\hline
\end{tabular}
\end{table}

\subsection{Experimental results of trajectory planning and control}

\begin{figure*}[]
    \centering
    \begin{minipage}{.29\linewidth}
            \includegraphics[width=\textwidth]{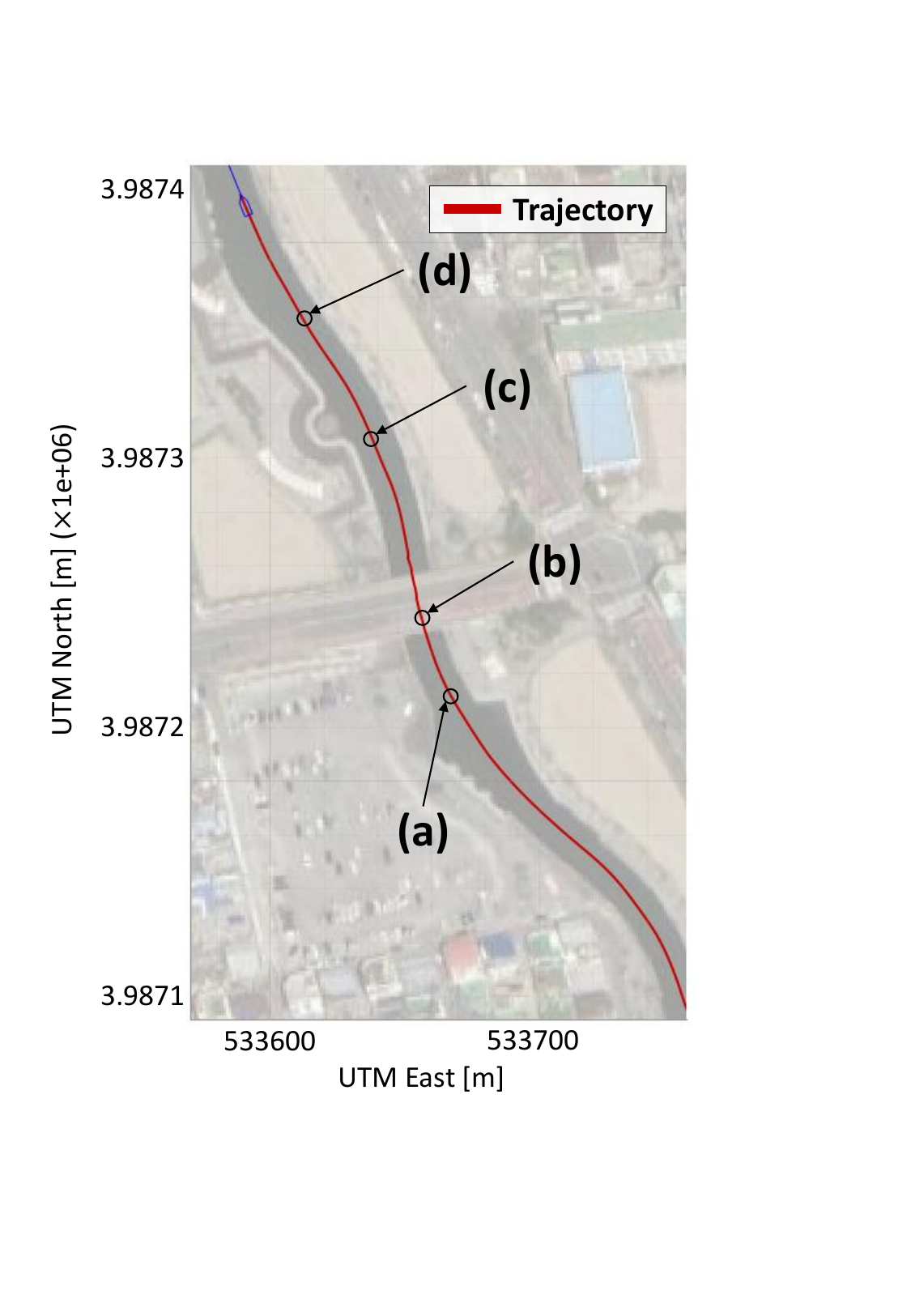}
            \captionsetup{labelformat=empty}
    \end{minipage}
    \begin{minipage}{.7\linewidth}
        \subfloat[][t = 294.4 sec]
            {\includegraphics[width=0.5\textwidth]{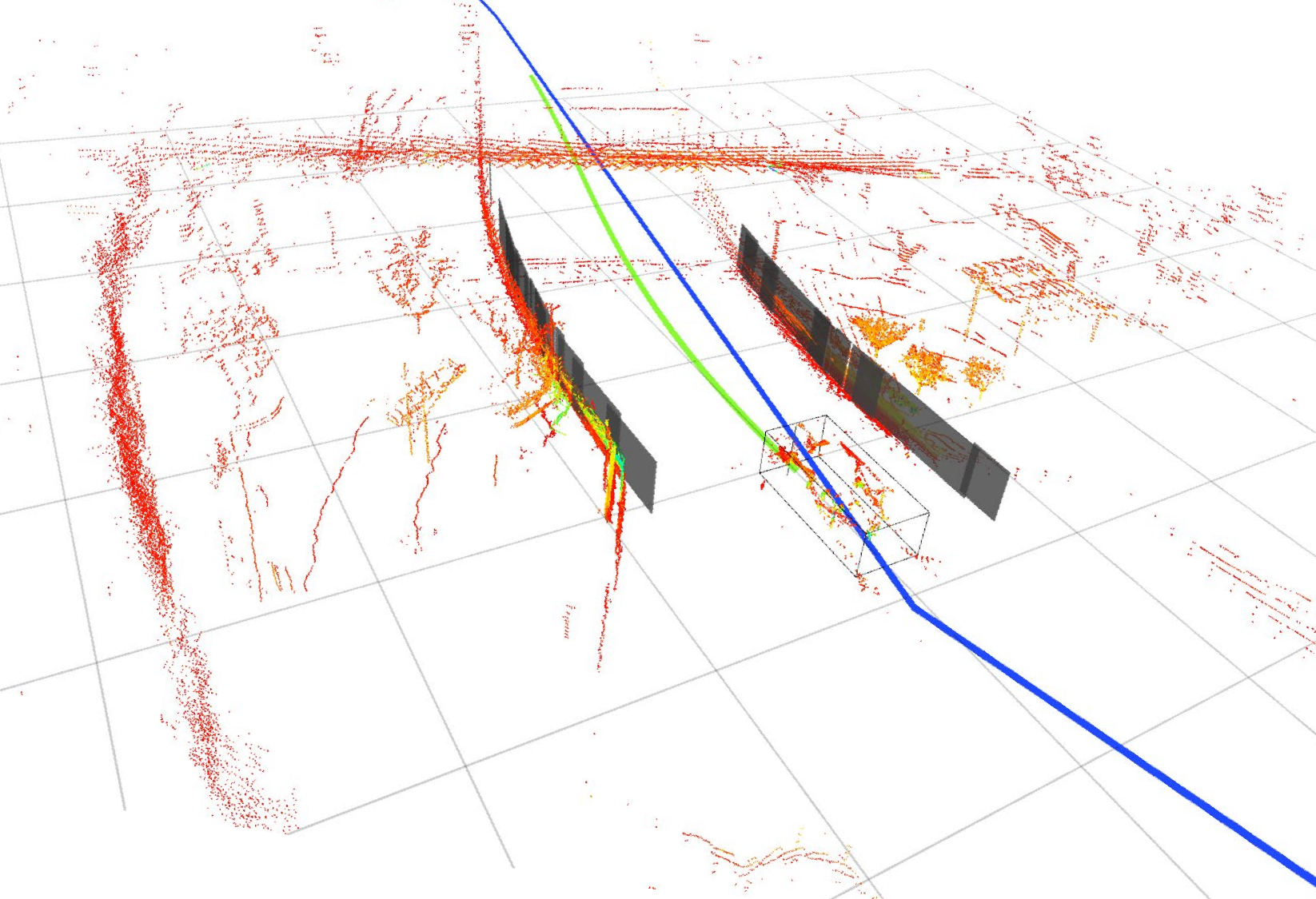}\label{fig:rviz1_1}}
        \subfloat[][t = 309.1 sec]
            {\includegraphics[width=0.5\textwidth]{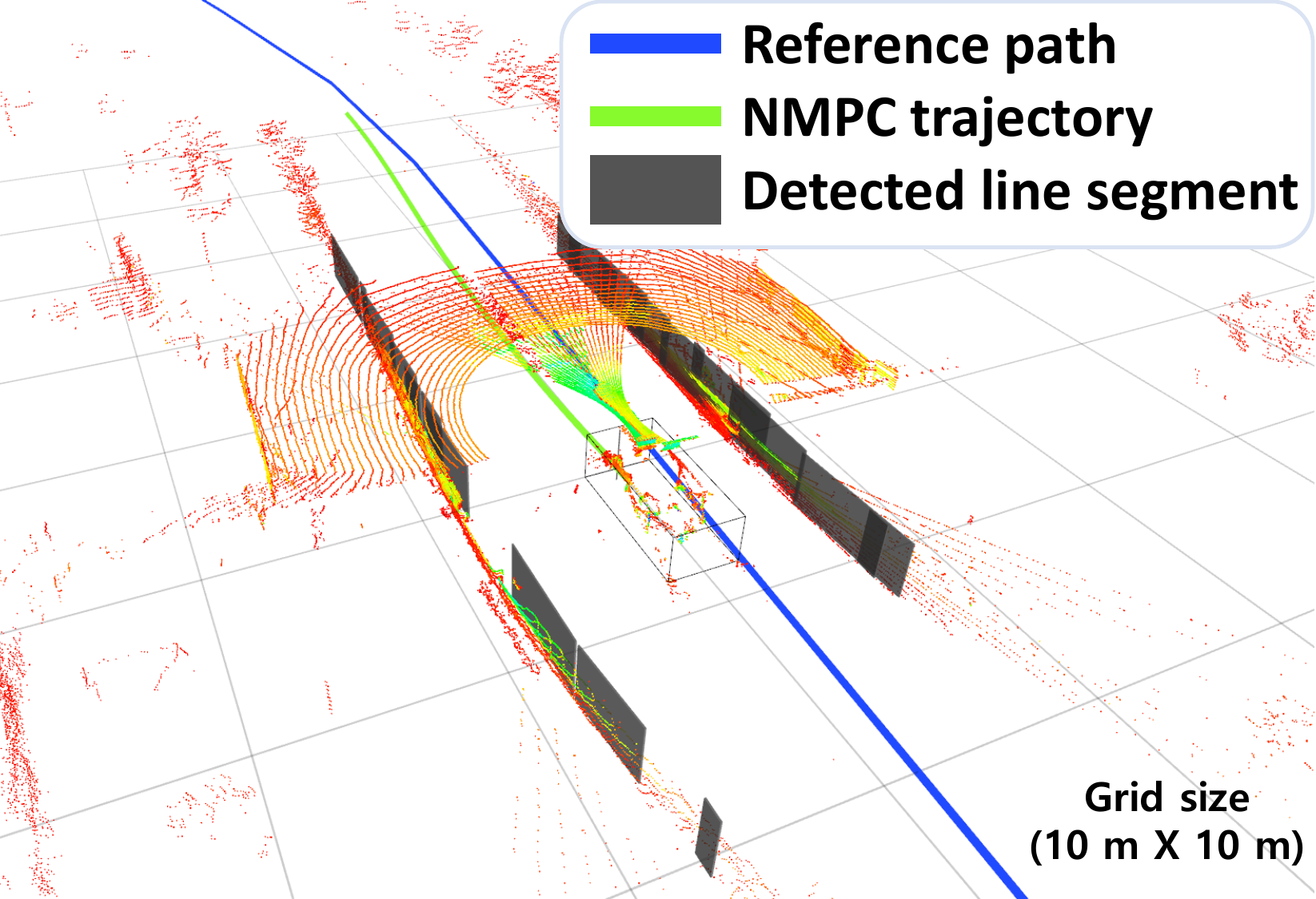}\label{fig:rviz1_2}}
            \par
        \subfloat[][t = 340.7 sec]
            {\includegraphics[width=0.5\textwidth]{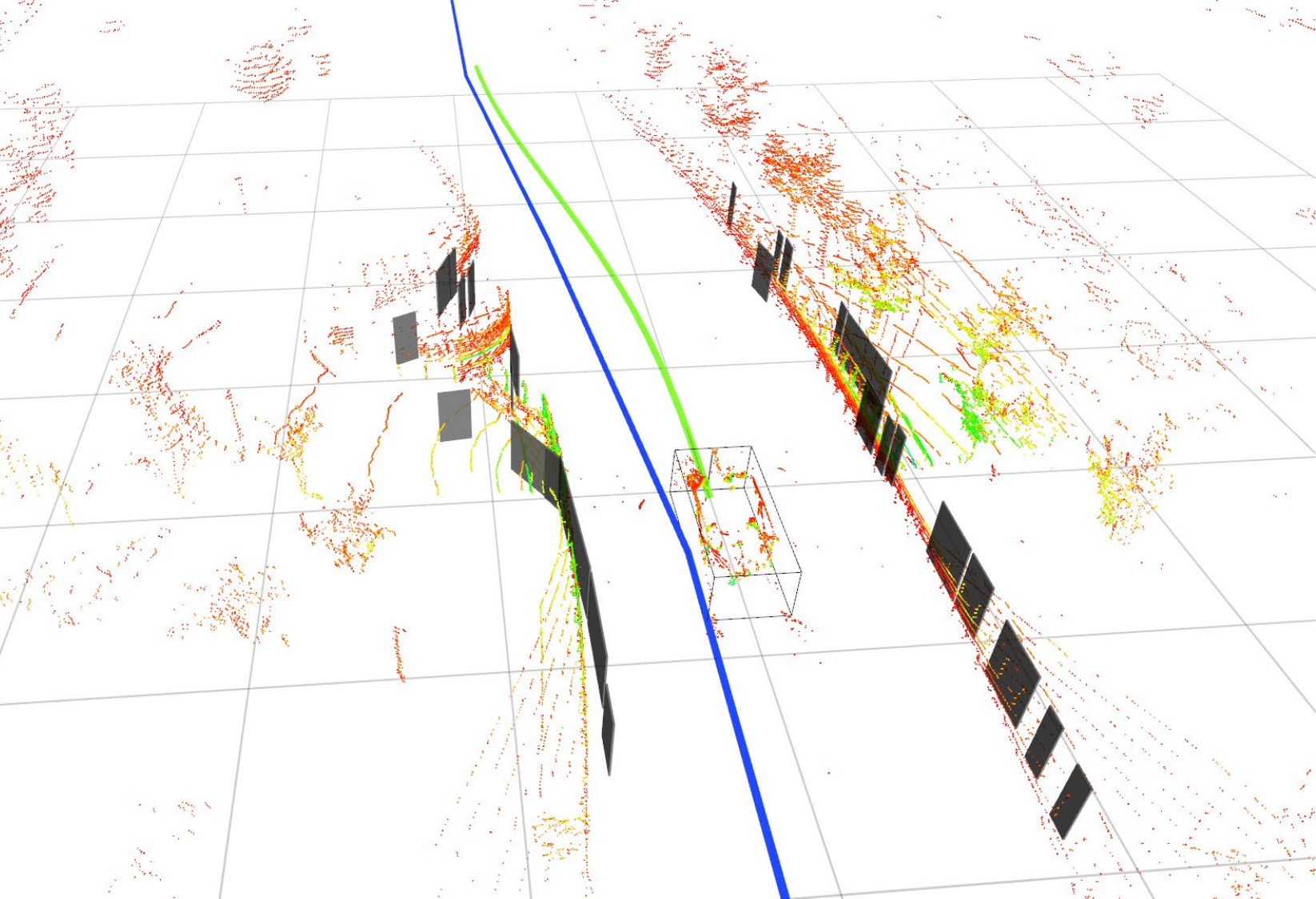}\label{fig:rviz1_3}}
        \subfloat[][t = 360.4 sec]
            {\includegraphics[width=0.5\textwidth]{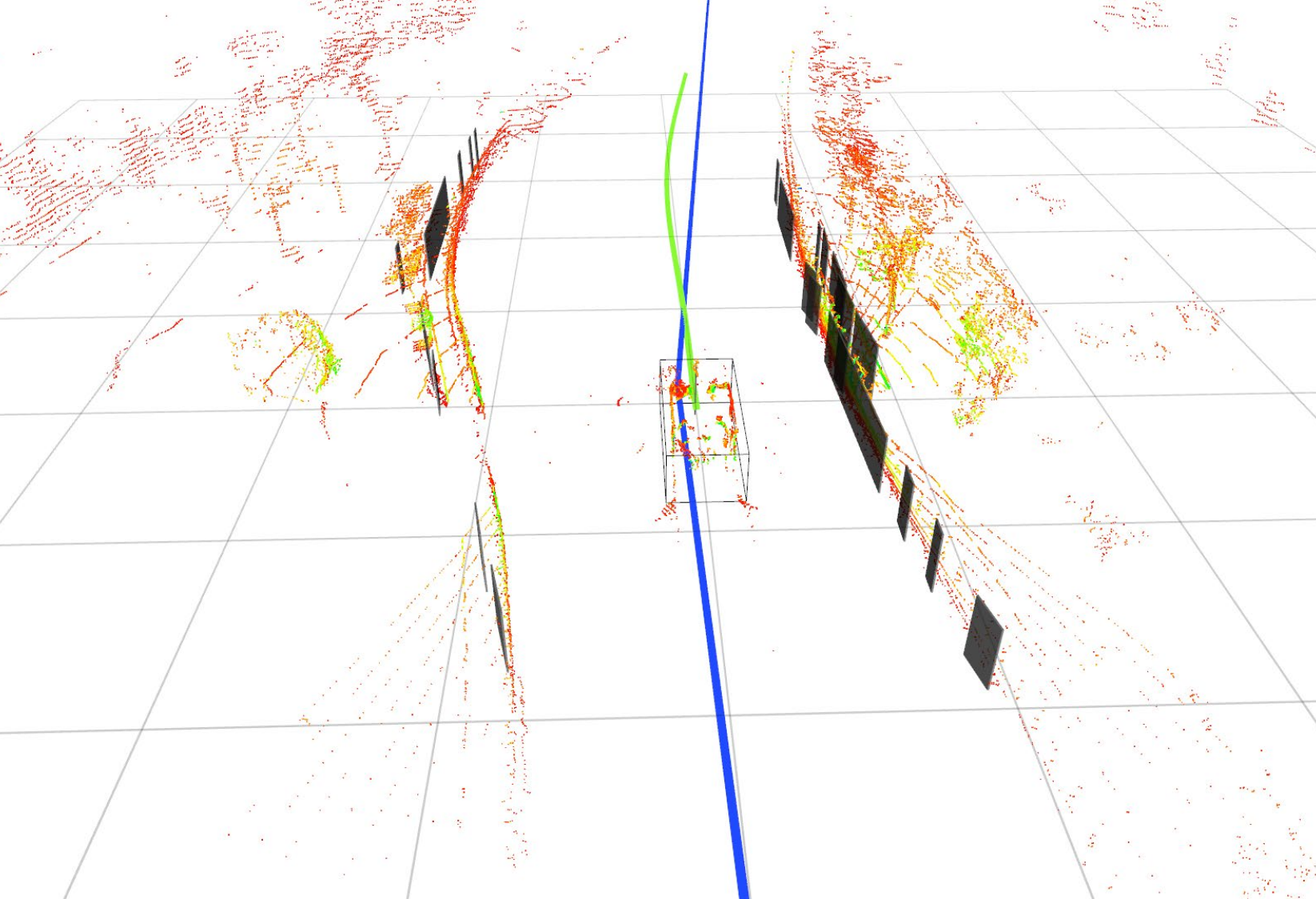}\label{fig:rviz1_4}}
    \end{minipage}
    \caption{Snapshot images when passing area \#1 during the experiment.
We visualized the point clouds, reference path (blue line), the results of the proposed NMPC (green line), and the detection algorithm (black patches). The grid size was 10 m. In the figure on the left, the locations of (a)-(d) are indicated on Google Maps.}
    \label{fig:Area1_Results}
\end{figure*}

\begin{figure*}[]
    \centering
    \begin{minipage}{.29\linewidth}
            \includegraphics[width=\textwidth]{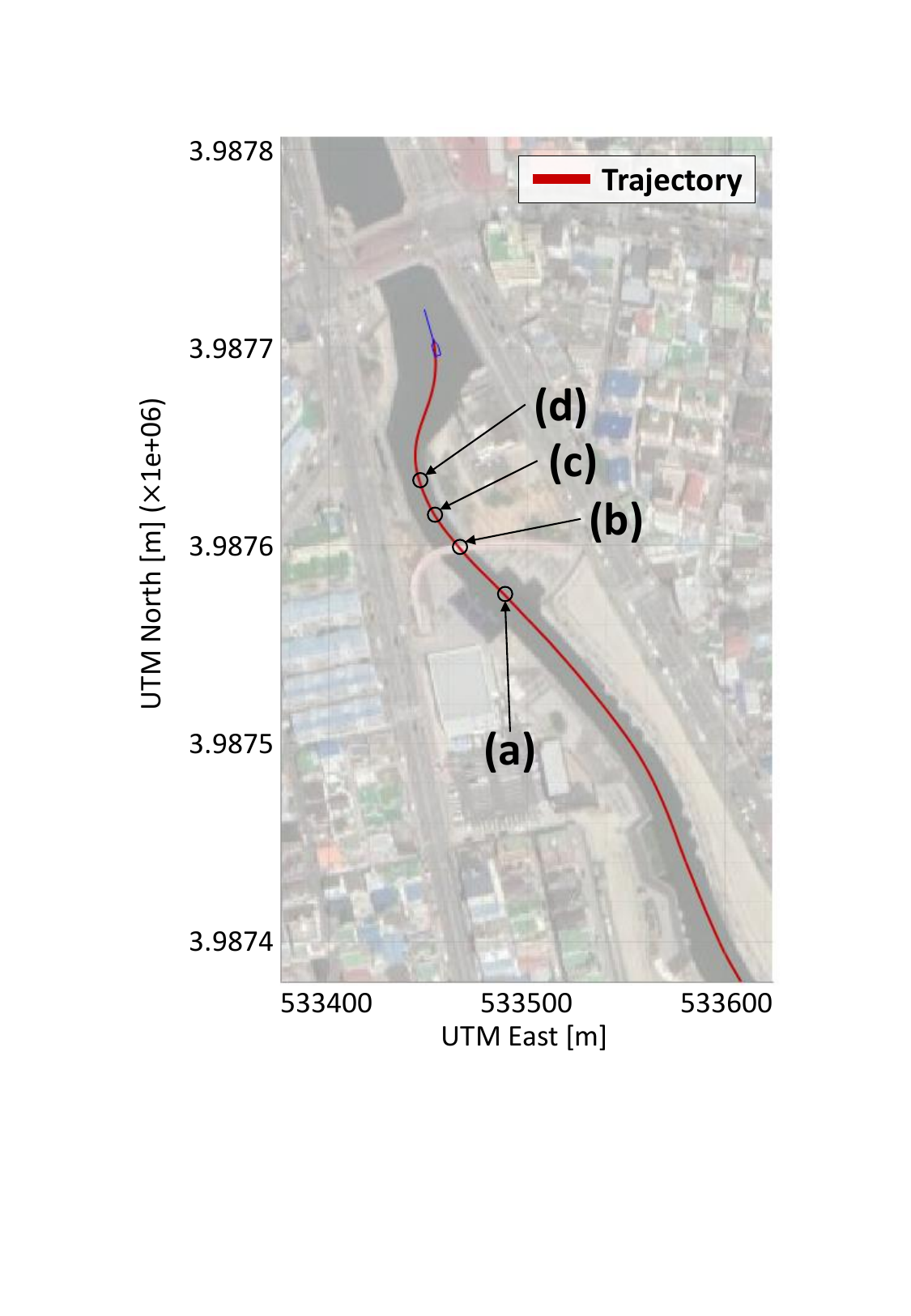}
    \end{minipage}
    \begin{minipage}{.7\linewidth}
        \subfloat[t = 466.9 sec]
            {\includegraphics[width=0.5\textwidth]{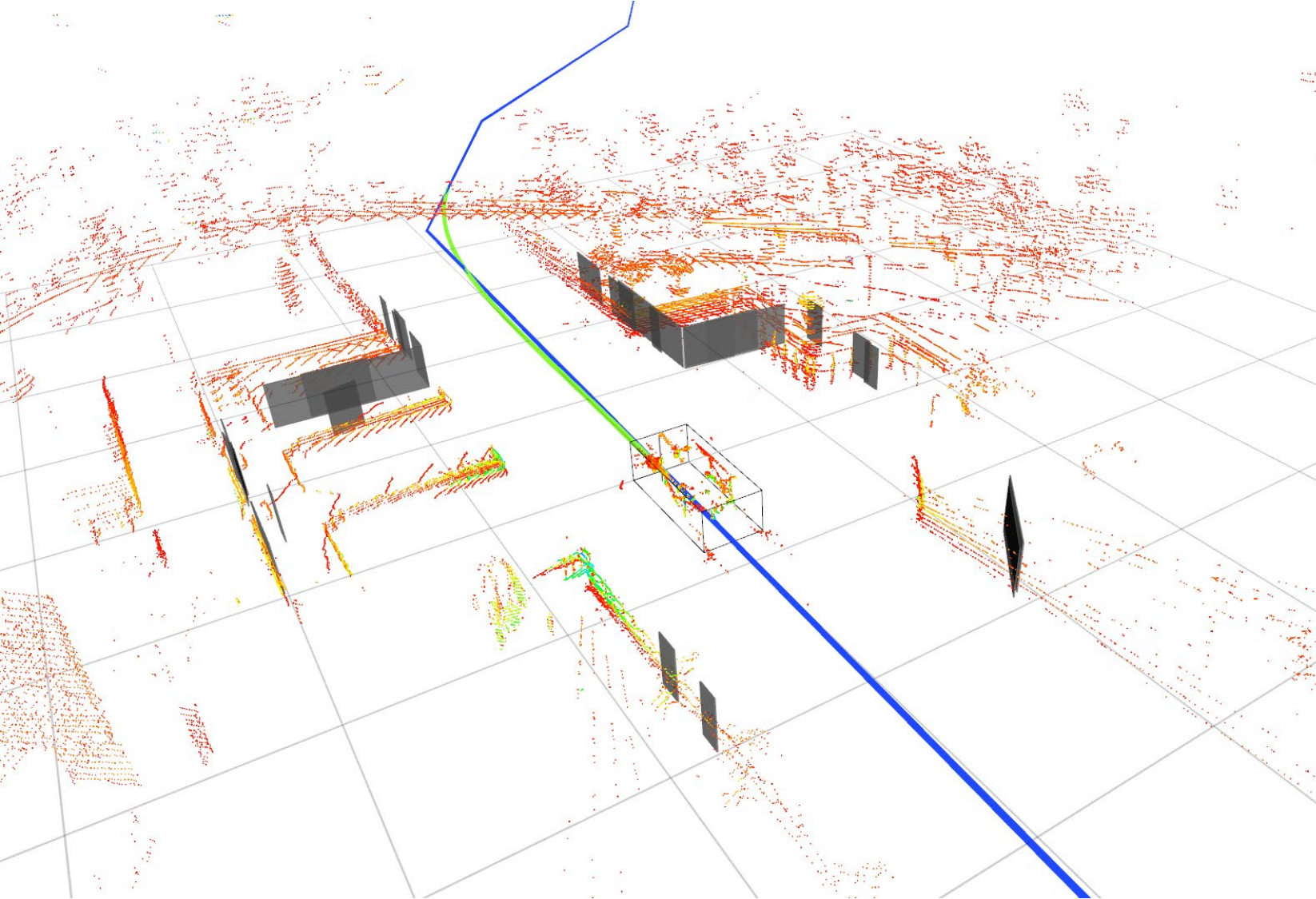}\label{fig:rviz2_1}}
        \subfloat[][t = 480.4 sec]
            {\includegraphics[width=0.5\textwidth]{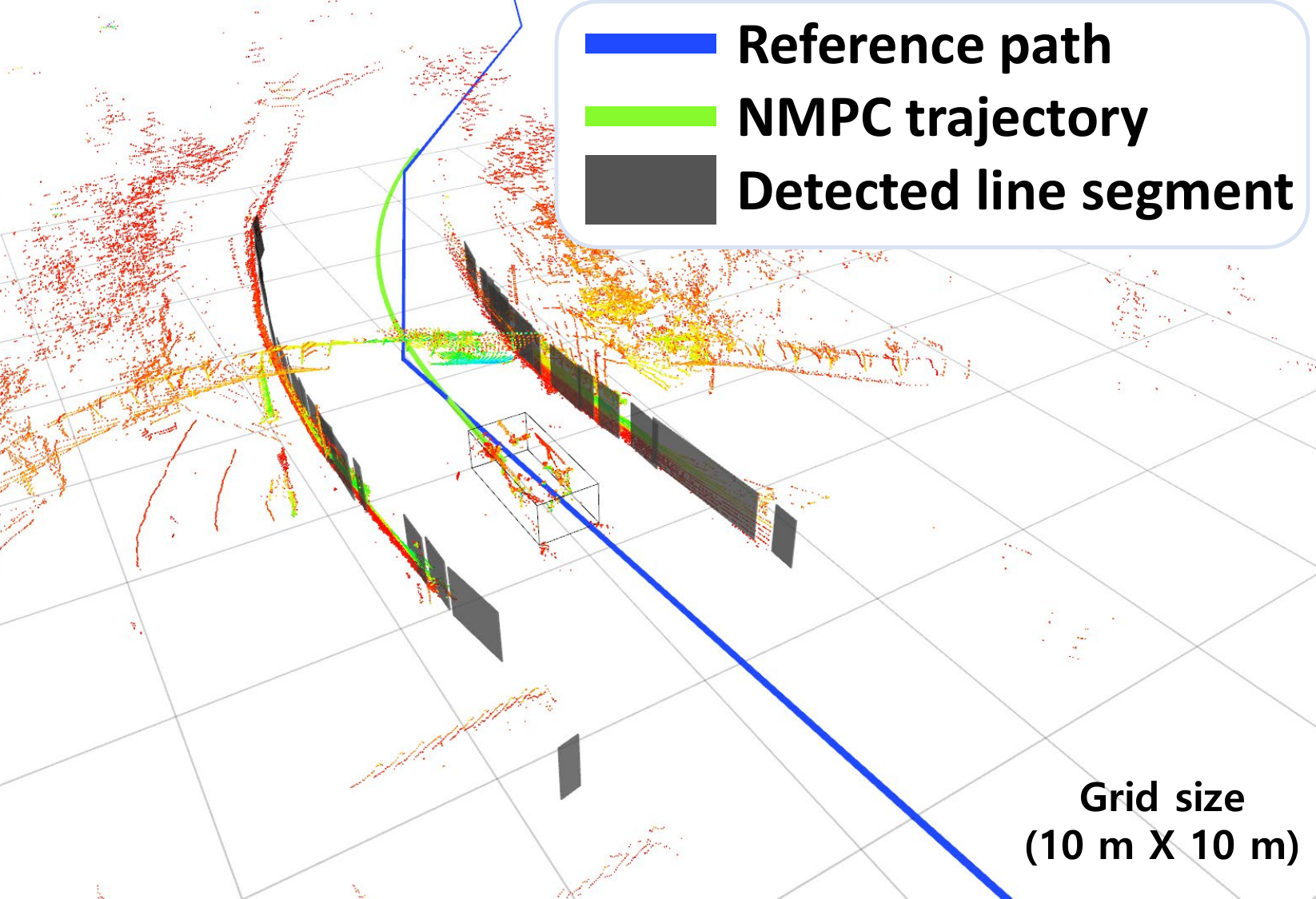}\label{fig:rviz2_2}}
            \par
        \subfloat[][t = 487.7 sec]
            {\includegraphics[width=0.5\textwidth]{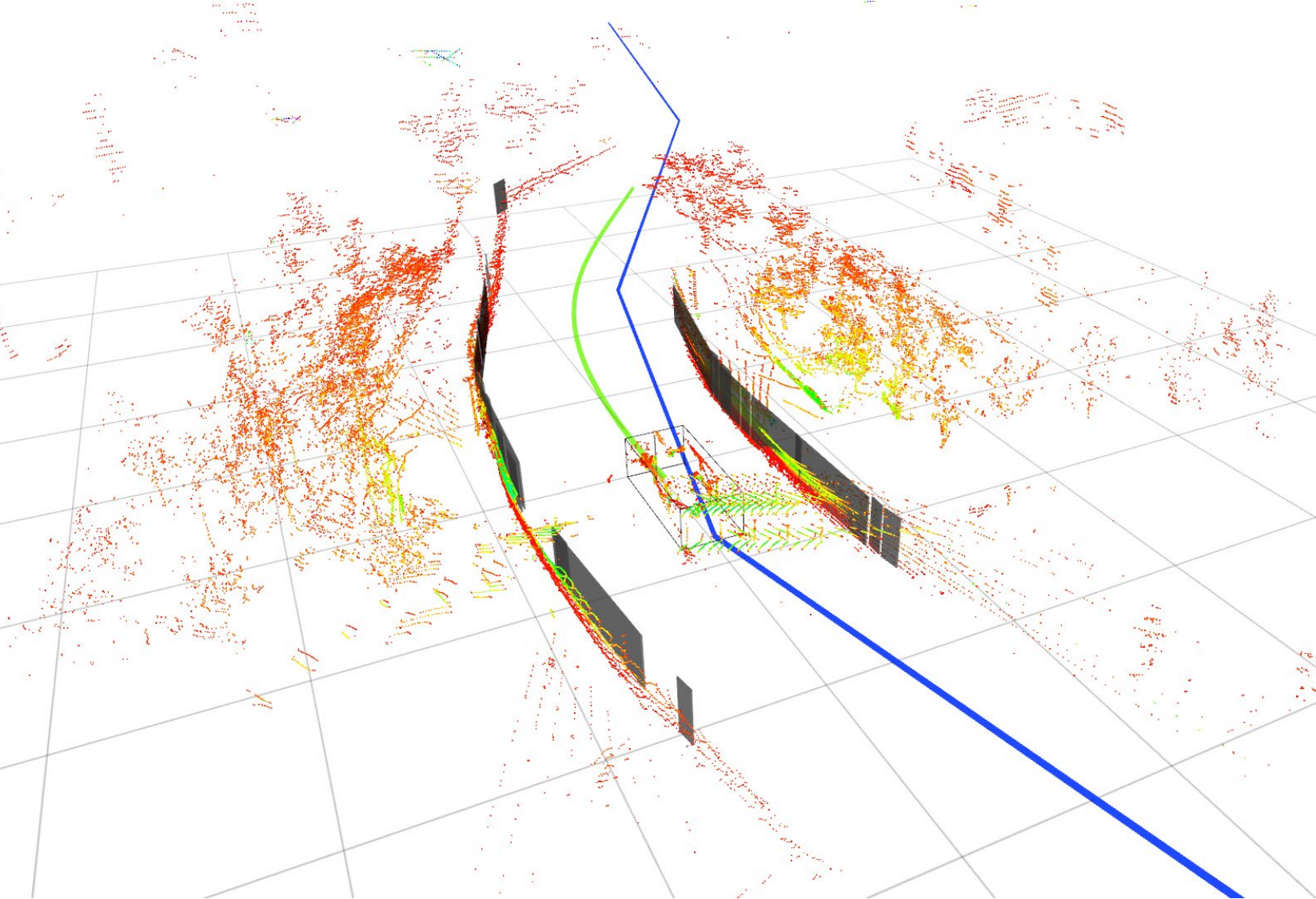}\label{fig:rviz2_3}}
        \subfloat[][t = 496.2 sec]
            {\includegraphics[width=0.5\textwidth]{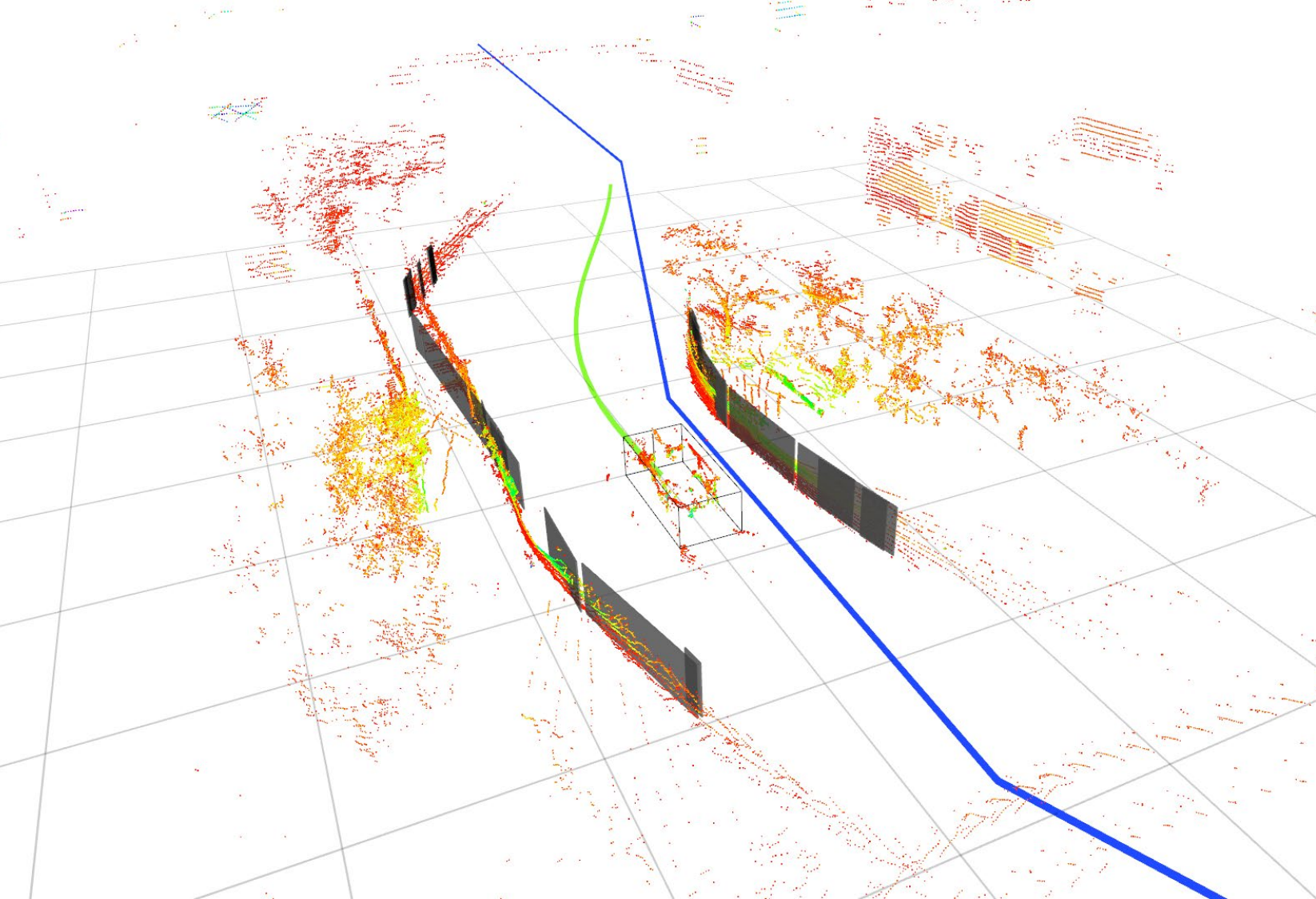}\label{fig:rviz2_4}}
    \end{minipage}
    \caption{Snapshot images when passing area \#2 during the experiment.
We visualized the point clouds, reference path (blue line), the results of the proposed NMPC (green line), and the detection algorithm (black patches). The grid size was 10 m. In the figure on the left, the locations of (a)-(d) are indicated on Google Maps.}
    \label{fig:Area2_Results}
\end{figure*}

\begin{figure*}[t]
\centering
    \includegraphics[width=\linewidth]{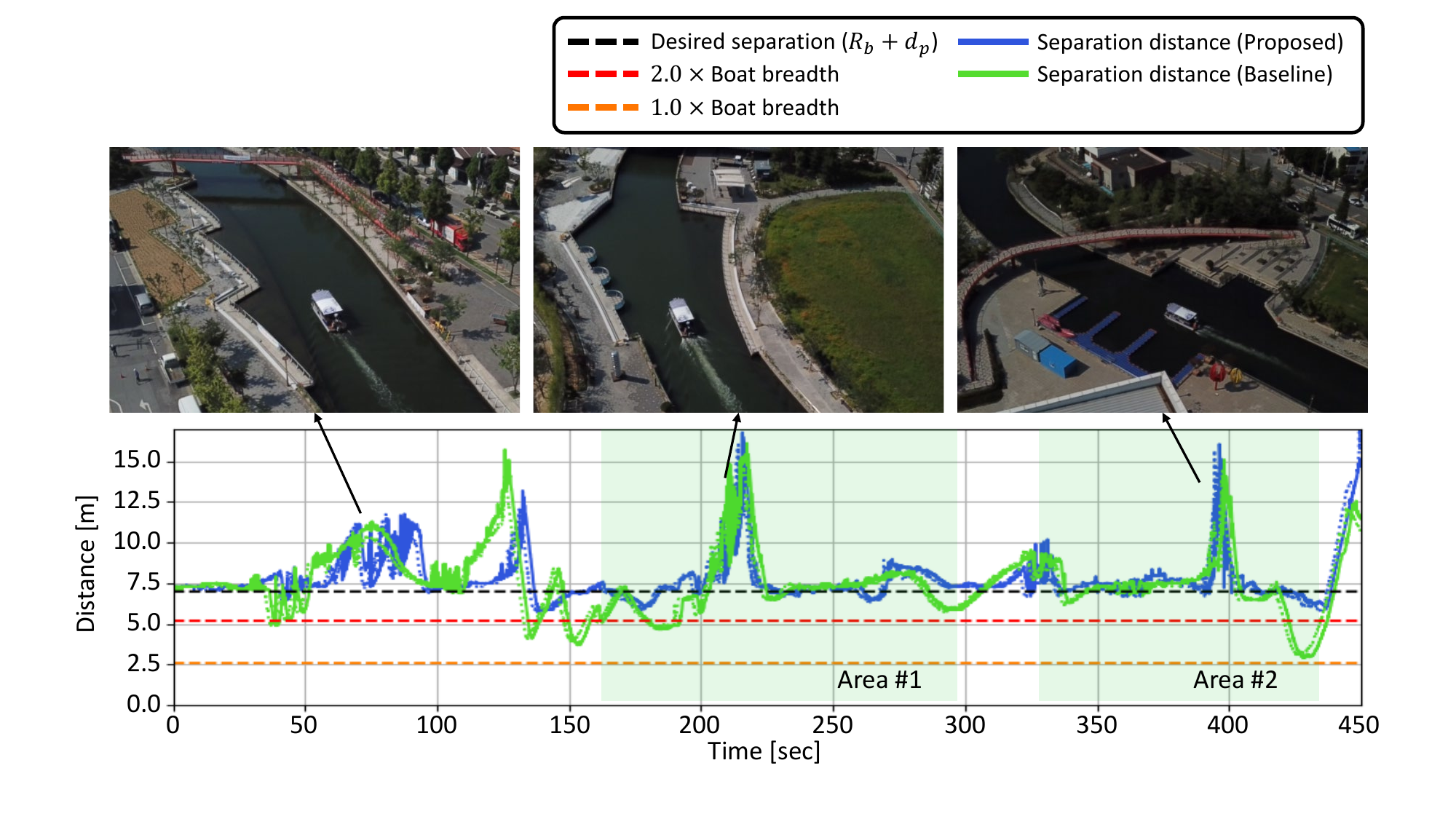}
    \caption{The blue and green lines represent the separation distance by proposed and baseline algorithms, and the dotted and solid lines indicate the distance from $p_b$ and $p_s$, respectively. The black dashed line indicates the desired separation ($R_{b}+d_p$) and the orange and red dashed lines represent the boat's breadth and twice of it, respectively. The green area represents the time spent in areas \#1 and \#2.
    The upper three photos show the moments the peak distance values are observed, which confirms that these sudden increases were due to the presence of some widened sections in the canal.
    }
    \label{fig:distance_Areas}
\end{figure*}
To validate the effectiveness of the proposed NMPC algorithm in a real-world environment, we conducted experiments in the Pohang Canal. This canal has an average width of 15 m and a length of 1 km. To define the obstacle avoidance constraints, we utilized LiDAR-based obstacle detection algorithms. As for the baseline algorithm, we chose the first baseline algorithm, as it showed better performance than the second baseline algorithm in the simulated environment, as described in Section~\ref{sec:sim}.

The trajectory results of the experiment are shown in Fig.~\ref{fig:trajectory}. For a detailed explanation, the results for areas \#1 and \#2 in Fig.\ref{fig:trajectory} are described.
Fig.~\ref{fig:Area1_Results} shows the snapshot images when passing area \#1. We plotted the trajectory on Google Maps and visualized the point cloud, reference path, and results of the proposed NMPC in Figs.~\ref{fig:rviz1_1}-\ref{fig:rviz1_4}. Fig.~\ref{fig:Area2_Results} shows the results obtained from the same method in area \#2.
The waypoint was set based on the cruise boat's route; however, due to a navigation error, it provided a dangerous path close to the obstacle structures, as shown in the figures.
The proposed method made it possible to detect obstacle structures nearby and maintain a predetermined separation distance to safely navigate through the canal. 

Figure.~\ref{fig:distance_Areas} shows the closest distance to the line segments detected by the LiDARs during the experiment. The two data represent the separation distance by the proposed and baseline algorithms.
The black dashed line indicates the desired separation between the obstacle and the boat: 7.0 m ($R_{b}+d_p$).
It can be seen that the constraint was satisfied in most instances throughout the experiment. 
The width of the canal varies along the path, but since the desired separation was 7.0 m, it inevitably violated the constraint in an area narrower than 14.0 m. In this area, the proposed algorithm tried to maintain an equal distance from both sides to minimize the cost for the slack variables, causing it to follow the center of the waterway. 
On the other hand, when the baseline algorithm was used, dangerous near-collision situations were observed. To quantify and compare control effort, the following equation was used to measure the amount of change in the control input during the experiments:
\begin{equation}
   J_c = \sum_{i=0}^{T}\mathbf{u}_i^\top R \mathbf{u}_i,
\end{equation}
where, the summation is taken over the duration of the experiment, which is denoted by $T$.
The values of the control effort metric for the proposed algorithm and the baseline algorithm were 925.63 and 1405.9, respectively. This result confirms that the proposed algorithm outperformed the baseline algorithm in terms of control efficiency.
More experimental results can be found in the supplementary video
(\url{https://youtu.be/p2MESqGvOSE}).

\section{Conclusion} \label{section5}
This paper proposed an NMPC-based optimal trajectory planning and tracking control algorithm for a cruise boat in a canal environment.
The nonlinear dynamics model of a boat was estimated by solving the nonlinear programming problem using experimental data from various test maneuvers, such as acceleration-deceleration and zigzag trials.
To avoid the obstacle structures in the canal environment, the information acquired through LiDARs was parameterized in the form of line segments.
Through consideration of the estimated vehicle dynamics model and obstacle detection results as constraints of the NMPC, obstacle avoidance, local trajectory planning, and tracking control could be performed in a single NMPC layer.
The proposed algorithm allowed for safe and successful autonomous navigation along the Pohang Canal and the practical feasibility of the proposed NMPC algorithm was verified.

\section*{Appendix} \label{section:appendix}
To conduct the simulation study and compare our approach with the second baseline algorithm \cite{shan2020receding}, we made certain modifications. Specifically, we employed a kinematic model to generate a smooth path, defined as follows:
\begin{equation}
    \dot{x} = u \cos \psi,\quad \dot{y} = u \sin \psi,\quad \dot{\psi} = r,
    \label{eq:baseline_kinematics}
\end{equation}
where the state and input vectors are defined as $\mathbf{x}_b = [x,y,\psi]^\top$, $\mathbf{u}_b = [u,r]^\top$.
And then, we formulated a two-point boundary value problem instead of using the sampling approach as follows:
\begin{equation}
    \label{eq:baseline}
    \min_{\mathbf{x}_b(\cdot),{\mathbf{u}_b}(\cdot)}\sum_{i=0}^{N_b-1}
     \mathbf{u}_{b,i}^\top P \mathbf{u}_{b,i},
\end{equation}
subject to 
\begin{subequations}
\begin{align}
    \mathbf{x}_{b,0} - \mathbf{x}_{i} &= 0,  \label{base:init} \\
    \mathbf{x}_{b,N_b} - \mathbf{x}_{f} &= 0,  \label{base:term} \\
    \mathbf{x}_{b,i+1} - f_{b,d}(\mathbf{x}_{b,i},\mathbf{u}_{b,i}) &= 0 , \ i = 0,\ldots,N_b-1, \label{base:dynamics} \\
    -[4,\ 0.1]^\top \leq \mathbf{u}_{b,i} &  \leq [4,\ 0.1]^\top , \ i=0,\ldots,N_b, \label{base:constraints1} \\        h(\mathbf{x}_{b,i}, L_j) & \leq 0, \ i=0,\ldots,N_b, \ j=0,\ldots,N_l, \label{base:constraints2} 
\end{align}
\end{subequations}
where $N_b$ is the prediction horizon, $P$ is a weight matrix, $\mathbf{x}_i$ and $\mathbf{x}_f$ are the initial and final state conditions, respectively. We set the final state as the $50$ m ahead point on the reference path, derived from $N_b = 50$, with 0.5 s sampling time and 2.0 m/s target speed.
$f_{b,d}$ in \eqref{base:dynamics} is a discretized model of \eqref{eq:baseline_kinematics}.
\eqref{base:constraints1} is an input saturation, which denotes the maximum speed and turn rate, and \eqref{base:constraints2} is an obstacle avoidance constraint same as \eqref{mpc:constraints2}.
Since \cite{shan2020receding} dealt with collision risk with the highest priority, we set it as a constraint here so that it can have a highest priority.
First, we computed the minimum cost path for heading using a weight matrix $P = \text{diag}([0,1])$, which resulted in a minimum cost of $J_1^*$. We then utilized a different weight matrix $P = \text{diag}([1,0])$ and added constraints to ensure that the total heading cost did not exceed $J_1^*$, as follows: 
\begin{equation}
\min_{\mathbf{x}_b(\cdot),{\mathbf{u}_b}(\cdot)}\sum_{i=0}^{N_b-1}\mathbf{u}_{b,i}^\top  P \mathbf{u}_{b,i} \leq J_1^*.
\end{equation}
This allowed us to reduce the distance cost while keeping the heading cost smaller than the previous minimum.
It is expected that the two-point boundary value problem will yield better performance than the sampling approach because it optimizes the path in the continuous space.

The formulated nonlinear program \eqref{eq:baseline} is solved using the interior point algorithm \cite{IPOPT} in the MATLAB environment along with the CasADi optimization library \cite{casadi}.

\bibliographystyle{IEEEtran}
\bibliography{IEEEabrv,main}
\begin{IEEEbiography}[{\includegraphics[width=1in,height=1.25in,clip,keepaspectratio]{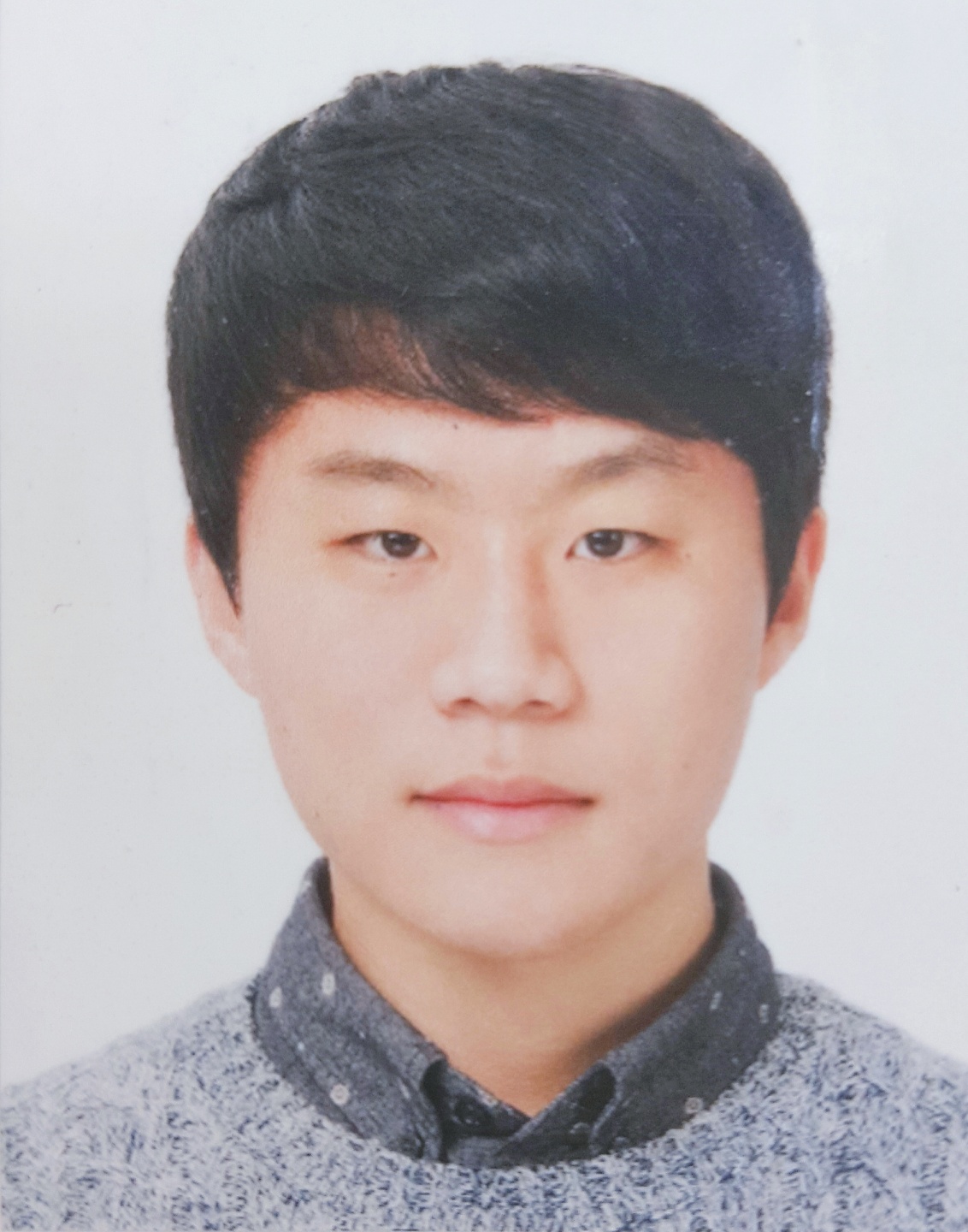}}]{Changyu Lee}
received the B.S. degree in Mathematics from Hanyang university, Seoul, South Korea in 2018, and the M.S. degree in Mechanical Engineering from Korea Advanced Institute of Science and Technology (KAIST), Daejeon, South Korea in 2020. He is currently working toward the Ph.D.degree in the Department of Mechanical Engineering at KAIST. His research interests include nonlinear control and model predictive control.
\end{IEEEbiography}
\begin{IEEEbiography}[{\includegraphics[width=1in,height=1.25in,clip,keepaspectratio]{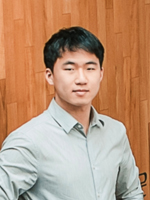}}]{Dongha Chung}
received B.S. and M.S. degrees in Mechanical Engineering from Korea Advanced Institute of Science and Technology (KAIST), Daejeon, South Korea in 2015 and 2017, respectively. He is currently working toward the Ph.D.degree in the Department of Mechanical Engineering at KAIST. His research interests include computer vision and visual/LiDAR simultaneous localization and mapping.
\end{IEEEbiography}
\begin{IEEEbiography}[{\includegraphics[width=1in,height=1.25in,clip,keepaspectratio]{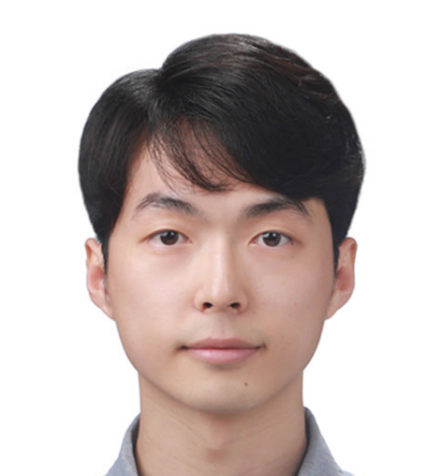}}]{Jonghwi Kim}
received B.S. and M.S. degrees in Mechanical Engineering from Korea Advanced Institute of Science and Technology (KAIST), Daejeon, South Korea in 2017 and 2019, respectively. He is currently working toward the Ph.D.degree in the Department of Mechanical Engineering at KAIST. His research interests include sensor fusion and vehicle localization.
\end{IEEEbiography}
\begin{IEEEbiography}[{\includegraphics[width=1in,height=1.25in,clip,keepaspectratio]{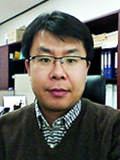}}]{Jinwhan Kim}
received B.S. and M.S. degrees in naval architecture and ocean engineering from Seoul National University, Seoul, South Korea, in 1993 and 1995, respectively, and the M.S. and Ph.D. degrees in aeronautics and astronautics from Stanford University, Stanford, CA, USA, in 2002 and 2007, respectively. 
He was a Full-Time Researcher with Korea Institute of Machinery and Materials and subsequently with Korea Ocean Research and Development Institute. He was a Research Scientist with Optimal Synthesis Inc., Los Altos, CA, USA. In 2010, he joined the Faculty with the Korea Advanced Institue of Science and Technology, Daejeon, South Korea. His research interests include robotics and the guidance, control, and estimation of dynamical systems.
\end{IEEEbiography}

\end{document}